\documentclass[10pt,journal,cspaper,compsoc]{IEEEtran}
\ifCLASSOPTIONcompsoc
\else
\fi
\ifCLASSINFOpdf
\else
\fi
\ifCLASSOPTIONcompsoc
\usepackage[normalsize,sf]{subfigure}
\else
\usepackage[footnotesize]{subfigure}
\fi

\ifCLASSOPTIONcompsoc
  \usepackage[caption=false,font=normalsize,labelfont=sf,textfont=sf]{subfig}
\else1
  \usepackage[caption=false,font=footnotesize]{subfig}
\fi
\usepackage{times}
\usepackage{epsfig}
\usepackage{graphicx}
\usepackage{amsmath}
\usepackage{amssymb}
\usepackage{multirow}
\usepackage[algoruled,vlined,linesnumbered]{algorithm2e}
\usepackage{rotating}
\usepackage{bbm}
\usepackage{color}
\usepackage{url}

\usepackage{algorithmic}

\DeclareMathAlphabet{\mathpzc}{OT1}{pzc}{m}{it} 
\DeclareMathOperator*{\argmax}{argmax}
\DeclareMathOperator*{\argmin}{argmin}

\newcommand{\eg}{e.g.\!}
\newcommand{\ie}{i.e.\!}

\begin{document}

%
\title{Learning from Multiple Sources for Video Summarisation}
%
%
%
%

\author{Xiatian~Zhu,~\IEEEmembership{Student Member,~IEEE,}
		Chen~Change~Loy,~\IEEEmembership{Member,~IEEE,}
        and~Shaogang~Gong
\IEEEcompsocitemizethanks{\IEEEcompsocthanksitem 
Xiatian Zhu and Shaogang Gong are with School of Electronic Engineering and Computer Science, Queen Mary University of London. 
\protect\\
E-mail: xiatian.zhu@qmul.ac.uk, s.gong@qmul.ac.uk 
\IEEEcompsocthanksitem Chen Change Loy is with Department of Information Engineering, 
The Chinese University of Hong Kong.
\protect\\
E-mail: ccloy@ie.cuhk.edu.hk }
\thanks{}}

%
%

\markboth{}
{X. Zhu \MakeLowercase{\textit{et al.}}: Multi-Source Video Summarisation}
%


\IEEEcompsoctitleabstractindextext{%
\begin{abstract}
Many visual surveillance tasks, e.g. video summarisation, is conventionally accomplished through analysing imagery-based features.
Relying solely on visual cues for public surveillance video
understanding is unreliable, since visual observations obtained from
public space CCTV video data are often not sufficiently trustworthy and events of interest can be subtle.
We believe that non-visual data sources such as weather reports and traffic sensory signals can be exploited to complement visual data for video content analysis and summarisation.
In this paper, we present a novel unsupervised framework to learn jointly from both visual and independently-drawn non-visual data sources for discovering meaningful latent structure of surveillance video data. 
In particular, we investigate ways to cope with discrepant dimension and
representation whilst associating these heterogeneous data sources, and
derive effective mechanism to tolerate with missing and incomplete data
from different sources.
We show that the proposed multi-source learning framework not only
achieves better video content clustering than
state-of-the-art methods, but also is capable of accurately inferring missing
non-visual semantics from previously-unseen videos. 
In addition, a comprehensive user study is conducted to validate the
quality of video summarisation generated using the proposed multi-source model.

\end{abstract}


\begin{keywords}
Multi-source data, heterogeneous data, visual surveillance,
clustering, event recognition, video summarisation.
\end{keywords}}

\maketitle

\IEEEdisplaynotcompsoctitleabstractindextext

%
\IEEEpeerreviewmaketitle

\label{sec:introduction}
Visual features and descriptors are often carefully designed
and exploited as the sole input for surveillance video content
analysis and summarisation.
For instance, optical or particle flow is typically employed in activity modelling~\cite{hospedales2011identifying,wang2009unsupervised,wu2010chaotic}, foreground pixel feature is used for multi-camera video understanding~\cite{loy2012incremental}, space-time image gradient is adopted for crowd analysis~\cite{kratz2012going}, and mixture of dynamic textures is used for video segmentation~\cite{chan2008modeling} and anomaly detection~\cite{li2013anomaly}.

A critical task in visual surveillance is to automatically make sense
of massive amount of video data by summarising its content using
higher-level intrinsic physical events\footnote{Spatiotemporal combinations of human activity or interaction patterns, e.g. gathering, or environmental state changes, e.g. raining.} beyond
low-level key-frame visual feature statistics and/or object detection counts.
In most contemporary techniques, low-level imagery visual cues are typically
exploited as the only information source for video
summarisation~\cite{KangCVPR2006,PritchPAMI2008,SFengCVPR2012,LeeCVPR2012,Lu_2013_CVPR}.  
On the other hand, in complex and cluttered public scenes there are
intrinsically more interesting and salient higher-level events that
can provide more meaningful and concise summarisation of the video
data. However, such events may not be visually well-defined (easily
detectable) nor detected reliably by visual cues alone. In particular,
surveillance visual data from public spaces is often
inaccurate and/or incomplete due to uncontrollable sources of variation, changes in
illumination, occlusion, and background clutters~\cite{gong2011security}.


\begin{figure*}[t]
\centering
\includegraphics[width=1\linewidth]{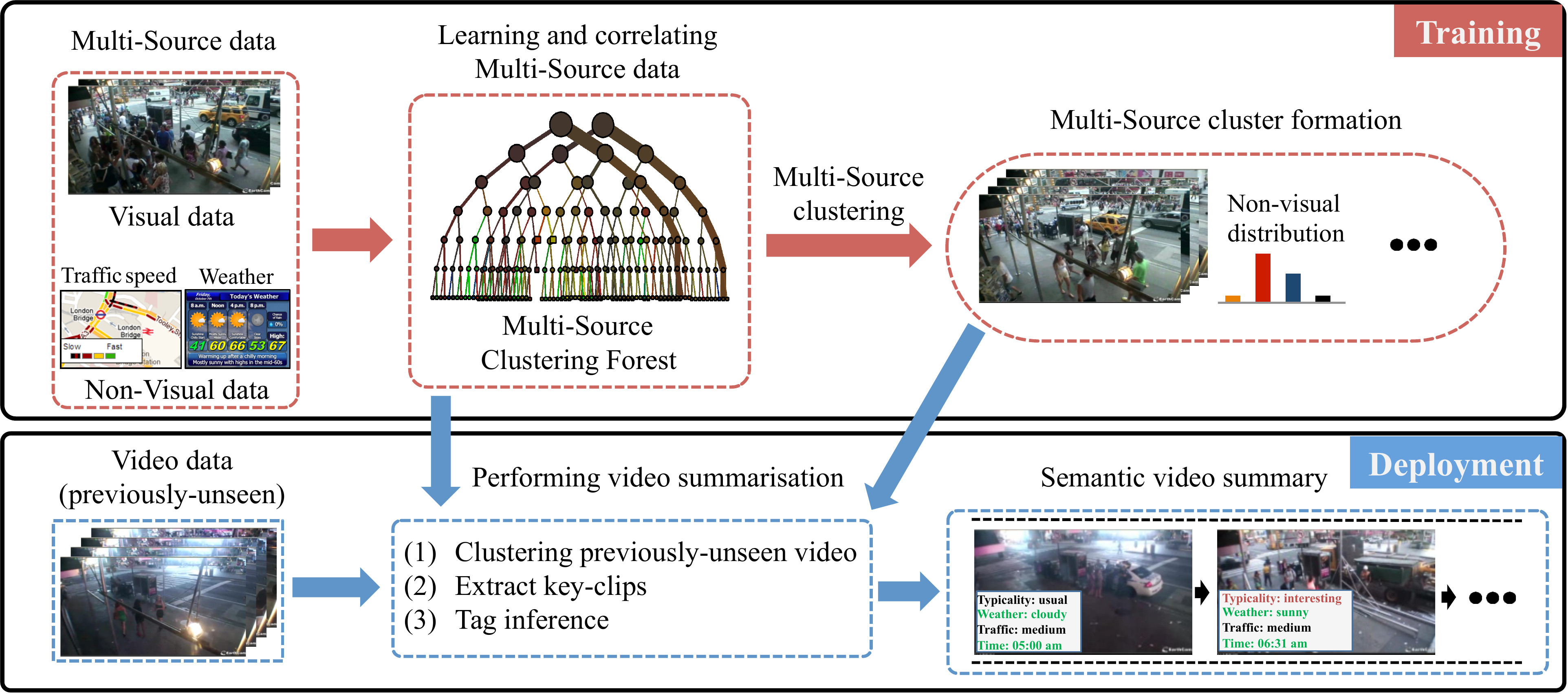}
\caption{\footnotesize The overview of the proposed multi-source driven video summarisation framework.
We consider a novel setting where multiple
  heterogeneous sources are present during the model training
  stage. The proposed Multi-Source Clustering Forest discovers and exploits latent
  correlations among heterogeneous visual and non-visual data sources
  both of which can be inaccurate and not trustworthy.
  In deployment, our model uncovers visual content structures and infer semantic tags 
  on previously-unseen video data for video summarisation.}
\label{fig:overview}
\end{figure*}

In this study, we wish to exploit non-visual auxiliary information to complement the unilateral perspective from visual observations. 
Examples of non-visual sources include weather report, GPS-based traffic data,
geo-location data, textual data from social networks, and on-line event
schedules. 
The auxiliary data sources are beneficial to visual data modelling 
because despite that visual and non-visual data may have very different 
characteristics and are of different natures, they depict the common physical
phenomenon in a scene.
They are
intrinsically correlated, although may be mostly indirect in some
latent spaces. 
Effectively discovering and exploiting such a latent correlation
space can facilitate the underlying data structure discovery and 
bridge the semantic gap between low-level visual features
and high-level semantical interpretation.

\vspace{0.1cm}
\noindent \textbf{Challenges} -
Nevertheless, it is non-trivial to formulate a framework that exploits both visual and non-visual data for video content analysis and summarisation, both algorithmically and in practice.

Algorithmically, unsupervised mining of latent correlation and interaction between
heterogeneous data sources faces a number of challenges:
(1) Disparate sources significantly differ in representation (continuous or categorical), and largely vary in scale and covariance\footnote{Also known as the heteroscedasticity problem~\cite{Duin_PAMI04}.}.
In addition, the dimension of visual sources often exceeds that of
non-visual information to a great extent ($>$2000 visual dimensions vs. $<$10 non-visual dimensions). Owing to this dimensionality discrepancy problem, a straightforward concatenation of features will result in a
representation unfavourably inclined towards the imagery data.
(2) Both visual and non-visual data in isolation can be inaccurate and
incomplete.

In practice, auxiliary data sources, e.g. weather, traffic reports, and event time tables, may be rather unreliable in availability. Specifically, the reports may not be released on-the-fly at a synchronised time stamp with the surveillance video stream. In addition, existing video control rooms may not necessarily have direct access to these sources. 
This renders models that expect complete visual and non-visual information during deployment impractical.


\vspace{0.1cm}
\noindent \textbf{Our solution} - In this study, we address this multi-source learning problem in the context of video summarisation, conventionally based on visual feature analysis and object
detection/segmentation.
In particular, we formulate a novel framework that is capable of performing joint learning given heterogeneous multi-sources (Fig.~\ref{fig:overview}).
We consider visual data as the \textit{main source} and non-visual data as the \textit{auxiliary sources}, since we believe visual information still plays the main role in video content analysis.
During training, we assume the access to both visual and non-visual data. 
The model performs multi-source data clustering and discovers a set of visual clusters tagged along with non-visual data distribution, e.g. different weathers and traffic speeds. 
We term the model as \textit{multi-source model}. 
During the deployment stage, we only assume the availability of previously-unseen video data since non-visual data may not be accessible due to the aforementioned limitations.
Since the learned model has already captured the latent structure of heterogeneous types of data sources, the model can be used for semantic video clustering and non-visual tag inference on previously-unseen video sequence, even without the non-visual data. Subsequently, key clips are automatically selected from the discovered clusters. The final summary video can be produced by chronologically compositing these key clips enriched by the inferred tags.

\vspace{0.1cm}
\noindent \textbf{Contributions} -
The main contributions of this work are:
\begin{enumerate}
\item 
We propose a unified multi-source learning framework capable of
  discovering semantic structures of video content collectively from
  heterogeneous visual and non-visual data. 
This is made possible by formulating a novel Multi-Source Clustering
Forest (MSC-Forest) that
seamlessly handles multi-heterogeneous data sources dissimilar in representation,
distribution, and dimension. 
Although both visual and non-visual data in isolation can be
inaccurate and incomplete, our model is capable of uncovering and subsequently exploiting the shared latent correlation for better data structure discovery.

\item 
The model is novel in its ability to 
accommodate partial or completely missing non-visual sources.
In particular, 
we introduce a joint information gain function 
that is capable of dynamically adapting to arbitrary amount of missing non-visual information during model learning. 
In model deployment, only visual input is required for inferring missing non-visual semantics.
\end{enumerate}

Extensive comparative evaluations are conducted on two public surveillance videos 
captured from both indoor and outdoor environments. 
Comparative results show that the proposed model not only outperforms the state-of-the-art methods~\cite{Huang2012,criminisi2012decision} for video content
clustering and structure discovery, but also is more superior in predicting
non-visual tags for previously-unseen videos.
The robustness of the proposed model is further validated by a user study on video summary quality.

\section{Related Work}
\label{sec:related_work}
\noindent \textbf{Multi-modality learning} - 
There exist studies that exploit different sensory or information modalities 
from a single source for data structure mining.
For example, 
Cai et al.~\cite{cai2011heterogeneous} propose to perform multi-modal image clustering by
learning a commonly shared graph-Laplacian matrix from different visual feature modalities.
Heer and Chi~\cite{heer2001identification} combine linearly individual similarity matrices derived from multi-modal webpages for web user grouping.
Karydis et al.~\cite{karydis2009tag} present a tensor based model to cluster music items with additional tags.
In terms of video analysis, the auditory channel and/or transcripts 
have been widely explored for detecting semantic concepts 
from multimedia videos~\cite{zhang2004semantic,fu2013learning},
summarising highlights in news and broadcast programs~\cite{Taskiran_MM06,Yihong_03},
or locating speakers~\cite{khalidov2011conjugate}.
User tags associated with web videos (e.g. YouTube) 
have also been utilised~\cite{ZWangCVPR2010,TodericiCVPR2010,wang2012event}.
In contrast, surveillance videos
captured from public spaces are typically without auditory
signals nor any synchronised transcripts and user tags available.
Instead, we wish to explore
alternative non-visual data drawn independently elsewhere from
multiple sources, with inherent challenges of being inaccurate and
incomplete, unsynchronised to and may also be in conflict with the
observed visual data.

\vspace{0.1cm}
\noindent \textbf{Multi-source learning} - 
%
%
An alternative multi-source learning mechanism can be
clustering ensemble~\cite{strehl2003cluster,topchy2005clustering} 
where a collection of clustering instances is generated and then aggregated into 
the final clustering solution. 
Typically only single data source is considered,
but it can be easily extended to handle multi-source data, e.g.
creating a respective clustering instance for each source.
Nonetheless, cross-source correlation is ignored 
since the clustering instances are separately formed and 
no interaction between them is involved.
%
A closer approach to ours is the 
Affinity Aggregation Spectral Clustering (AASC)~\cite{Huang2012}, 
which learns data structure from multiple types of homogeneous information (visual features only). Their method generates independently multiple affinity data matrices 
by exhaustive pairwise distance computation for every pair of samples in every data source. 
It suffers from unwieldy representation given high-dimensional data
inputs. Importantly, despite that it seeks for optimal weighted combination
of distinct affinity matrices, it does not consider correlation between
different sources in model learning, similar to 
clustering ensemble~\cite{strehl2003cluster,topchy2005clustering}.
Differing from the above models, our Multi-Source Clustering Forest overcomes these problems by generating a unified single affinity matrix that captures latent correlations among heterogeneous types of data sources.
Furthermore, our model has a unique advantage in handling missing non-visual data 
over~\cite{strehl2003cluster,topchy2005clustering,Huang2012}.

\vspace{0.1cm}
\noindent \textbf{Video summarisation} - 
Contemporary video summarisation methods can be broadly classified into two paradigms,
key-frame-based~\cite{LeeCVPR2012,WolfICASSP1996,ZhangPR1997,truong2007video,money2008video}
and object-based~\cite{PritchPAMI2008,SFengCVPR2012,PritchICCV2007} methods.
The key-frame-based approaches select representative key-frames 
by analysing low-level imagery properties, 
e.g. optical flow~\cite{WolfICASSP1996} or image differences~\cite{ZhangPR1997},
object's appearance and motion~\cite{LeeCVPR2012}, 
to form a storyboard of still images.
Object-based techniques~\cite{PritchPAMI2008,SFengCVPR2012}, 
on the other hand, rely on object segmentation and tracking 
to extract object-centric trajectories/tubes,
and compress those tubes to reduce spatiotemporal redundancy.

Both the above schemes utilise solely visual information and make
implicit assumptions about the completeness and accuracy of the visual
data available in extracting features or object-centered
representations. They are unsuitable nor scalable to complex
scenes where visual data are inherently incomplete and inaccurate,
mostly the case in surveillance videos.
Our work differs significantly to these studies in that we exploit not
only visual data without object tracking, 
but also non-visual sources as complementary information.
The summary generated by our approach is semantically enriched -- it
is labelled automatically with semantic tags, e.g. traffic condition,
weather, or event. All these tags are learned from heterogeneous
non-visual sources in an unsupervised manner during model training without any manual labels.




\vspace{0.1cm}
\noindent \textbf{Random forests} - 
Random forests~\cite{BreimanML01,criminisi2012decision} have proven as powerful models in the literature.
Different variants of random forests have been devised, either supervised~\cite{shotton2011real,GallYRGL11,schulter13d,Bosch07,caruana2008empirical}, or unsupervised~\cite{LiuCIKM2000,shi2006unsupervised,perbet2009random,moosmann2008randomized,zhu2014constructing}.
Supervised models are not suitable to our problem since we do not assume the availability of ground truth labels during model training.
Existing clustering forest models, on the other hand, assumes only homogeneous data sources such as pure imagery-based features.
No principled way of combining multiple heterogeneous and independent data sources in forest models is available. 


\section{Multi-Source Clustering}
\label{sec:method}

Video summarisation by content abstraction aims to generate a compact
summary composed of key/interesting content from a long previously-unseen video
for achieving efficient holistic understanding~\cite{truong2007video}.
A common way to establish a video summary is 
by extracting and then combining a set of key frames or shots.
These key contents are usually discovered and selected from 
clusters of video frames or clips~\cite{truong2007video}.

In this study, we follow the aforementioned approach 
but consider not only visual content of video, 
but also a large corpus of non-visual data collected 
from heterogeneous independent sources (Fig.~\ref{fig:data_structure_discovery}(a)).
Specifically, through learning latent structure of multi-source data (Fig.~\ref{fig:data_structure_discovery}(b-c)), 
we wish to make reference to and/or impose non-visual semantics 
directly into video clustering without any human manual annotation of video data (Fig.~\ref{fig:data_structure_discovery}(d)).
Formally, we consider the following different data sources that form a
multi-source input feature space:

\begin{figure}
\centering
\includegraphics[width=1\linewidth]{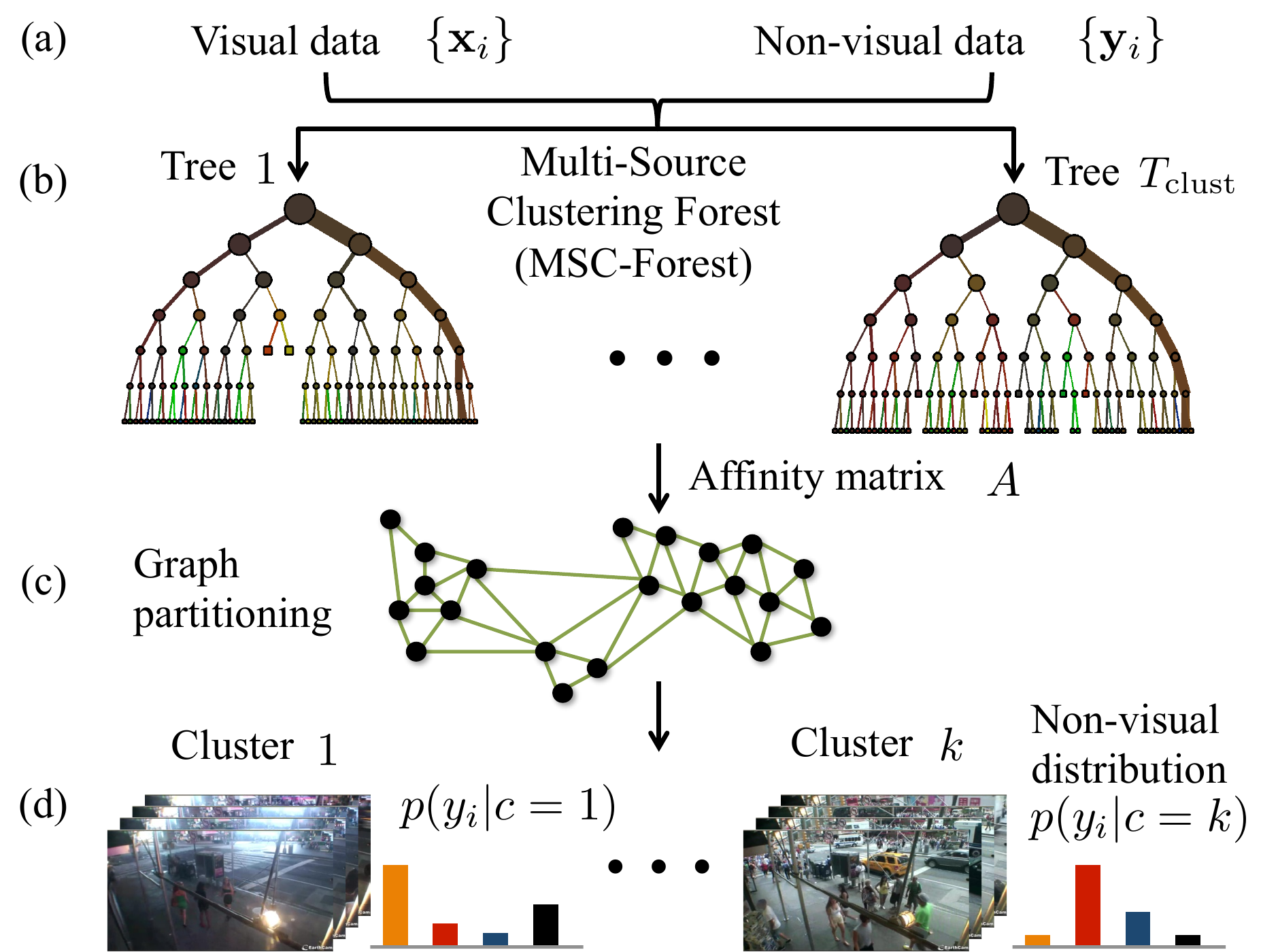}
\caption{Multi-source model training stage: The pipeline of performing multi-source clustering on visual and non-visual data with the proposed Multi-Source Clustering Forest (MSC-Forest).}
\label{fig:data_structure_discovery}
\end{figure}

\vspace{0.1cm}
\noindent \textbf{Visual features} - 
We segment a training video into $N$ either overlapping or non-overlapping clips, each of which has a duration of $T_\mathtt{clip}$ seconds. We then extract a $\mathit{d}$-dimensional visual descriptor from the $i$th video clip denoted by $ \mathbf{x}_i = \left( x_{i,1}, \dots, x_{i,\mathit{d}} \right) \in \mathbb{R}^{\mathit{d}}, i = 1, \dots, N $.

\vspace{0.1cm}
\noindent \textbf{Non-visual data} -  
Non-visual data are collected from heterogeneous independent sources. 
We collectively represent $\mathit{m}$ types of non-visual data associated with the $i$th clip as $\mathbf{y}_i = \left( y_{i,1},
\dots, y_{i,\mathit{m}} \right) \in \mathbb{R}^\mathit{m}$, $i = 1,
\dots, N $. 
Note that any (or all) dimension of $\mathbf{y}_i$ may be missing.

We aim at formulating a unified clustering model capable of coping with the few challenges as highlighted in Section~\ref{sec:introduction}.
The model needs be unsupervised since no ground truth is assumed.
To mitigate the heteroscedasticity and dimension discrepancy problems,
we require a model that can isolate the very different characteristics of visual and non-visual data, yet can still exploit their latent correlation in the clustering process.
To handle noisy data, feature selection is needed and necessary.

In light of the above demands, we choose to start with the
clustering random forest~\cite{BreimanML01,LiuCIKM2000,shi2006unsupervised} due to 
(1) unsupervised information gain optimisation thus requiring no ground truth labels;
(2) its flexible objective function for facilitating the modelling of multi-source data as well as the processing of missing data;
(3) and its implicit feature selection mechanism for handling noisy features.
Nevertheless, the conventional clustering forest is not well
suited to solve these challenges
since it expects a full concatenated representation 
as input during both model training and deployment. 
This does not conform to the assumption of only visual
data being available during model deployment for previously-unseen videos. 
Moreover, due to its uniform variable selection
mechanism~\cite{BreimanML01} 
(e.g. each feature dimension has the same probability to be selected as a candidate optimal splitting variable), 
there is no principled way to ensure balanced contribution from 
individual visual and non-visual sources in the node splitting process. 
To overcome these limitations, we 
propose a new \textit{Multi-Source Clustering Forest} (MSC-Forest) by
introducing a new objective function
allowing \textit{joint optimisation of individual information gains} of different sources. 
We first describe the conventional forests prior to detailing the proposed MSC-Forest.

\subsection{Conventional Random Forests}
\label{sec:method@RF_background}

\begin{figure*}
\centering
	\subfigure []
	{
		 \includegraphics[width=.23\linewidth]{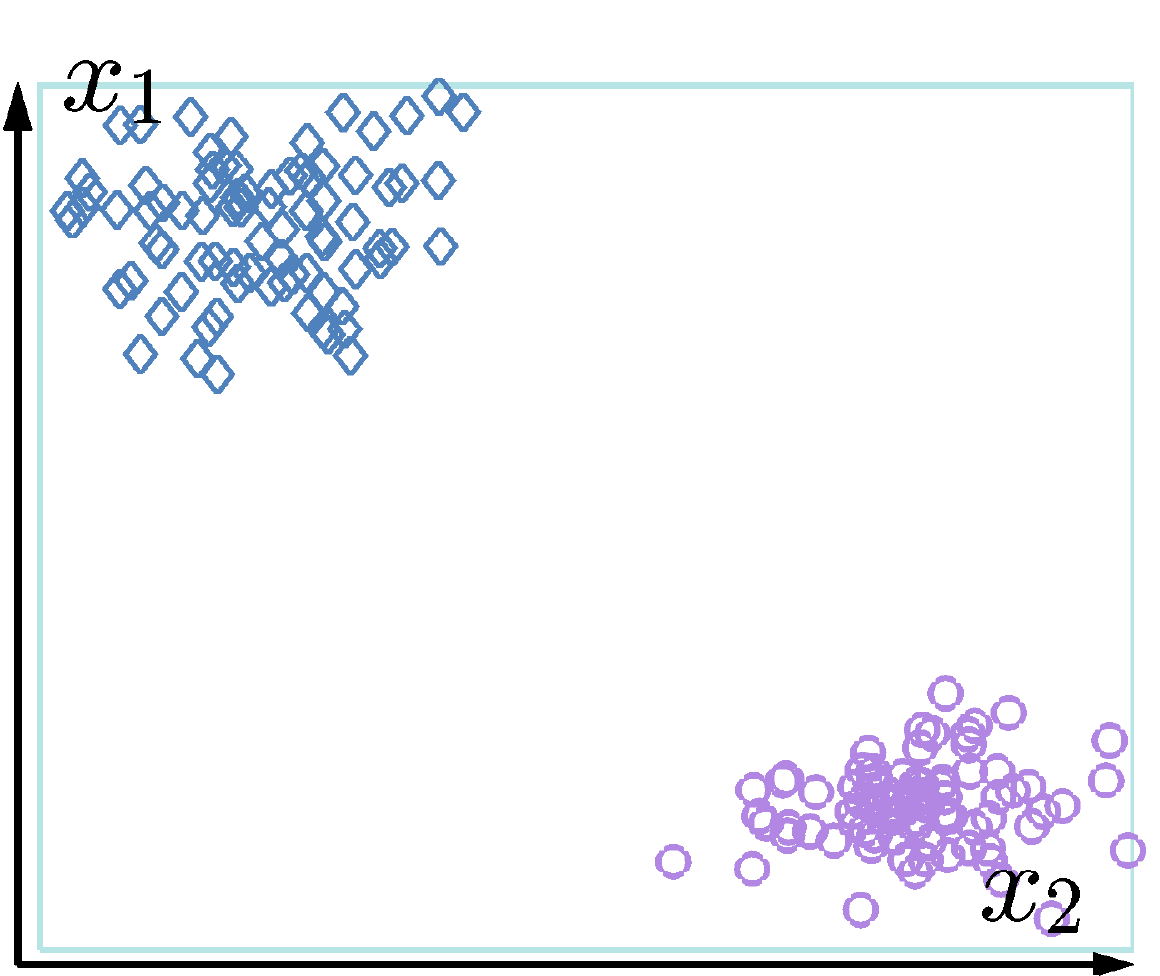}
	}
	\subfigure []
	{
		 \includegraphics[width=.23\linewidth]{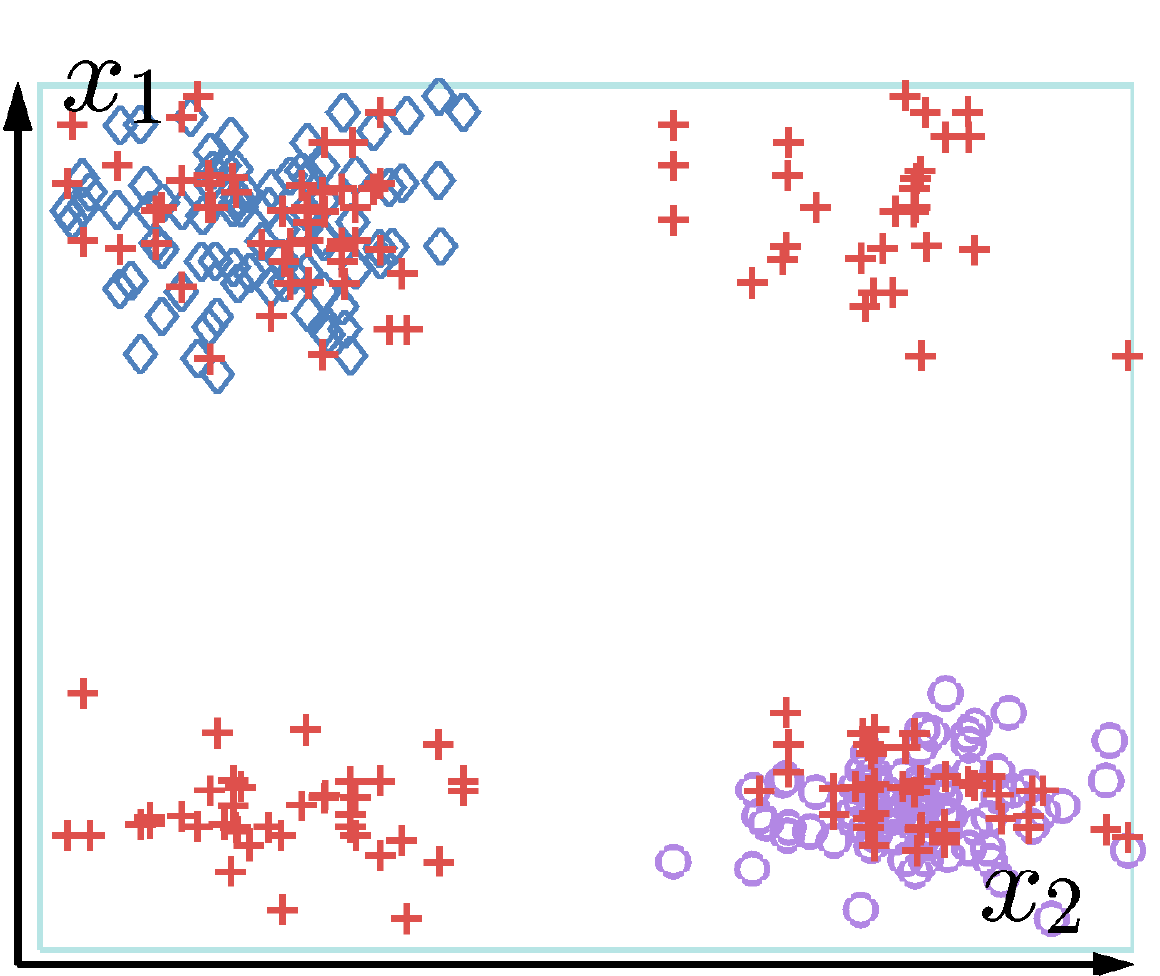}
	}
	\subfigure []
	{
		 \includegraphics[width=.23\linewidth]{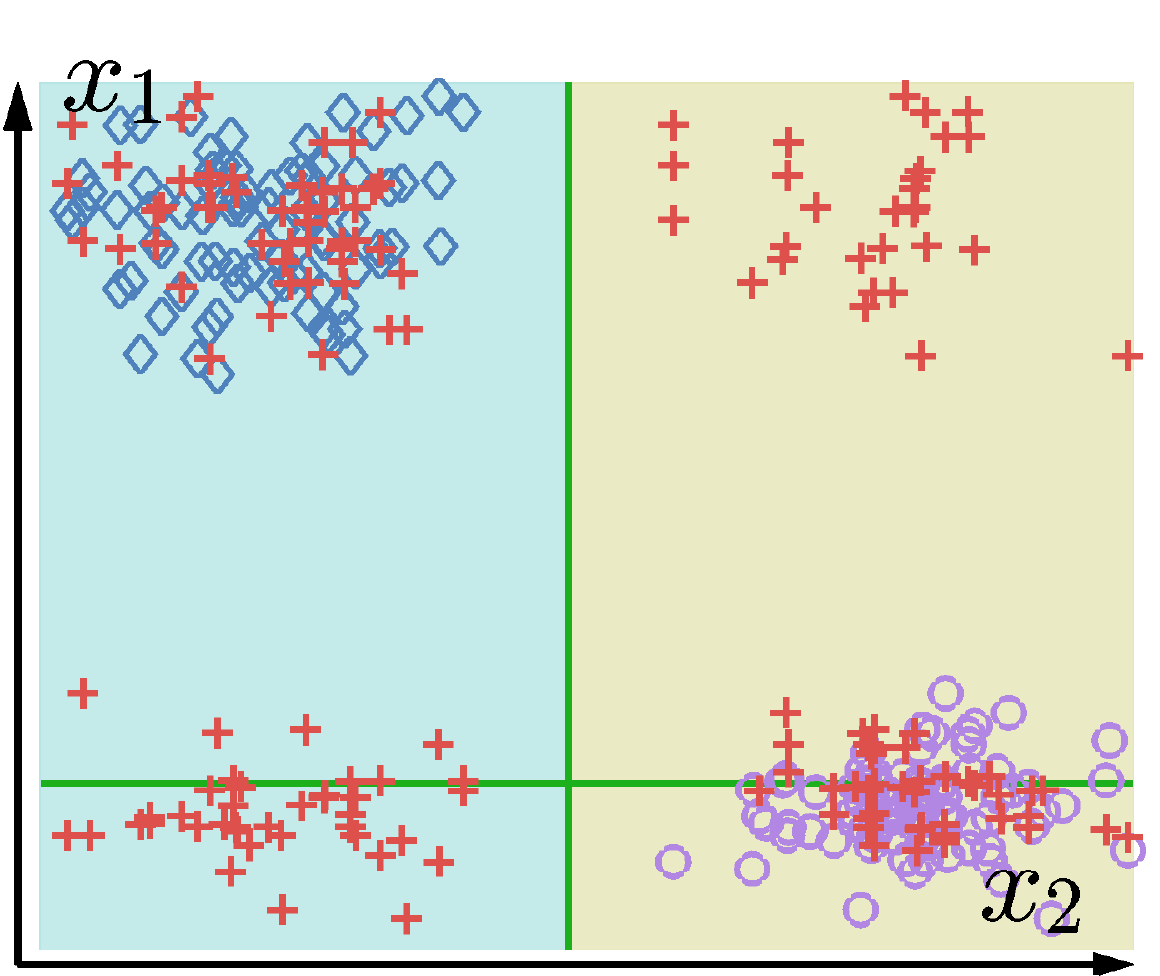}
	}
	\subfigure []
	{
		 \includegraphics[width=.23\linewidth]{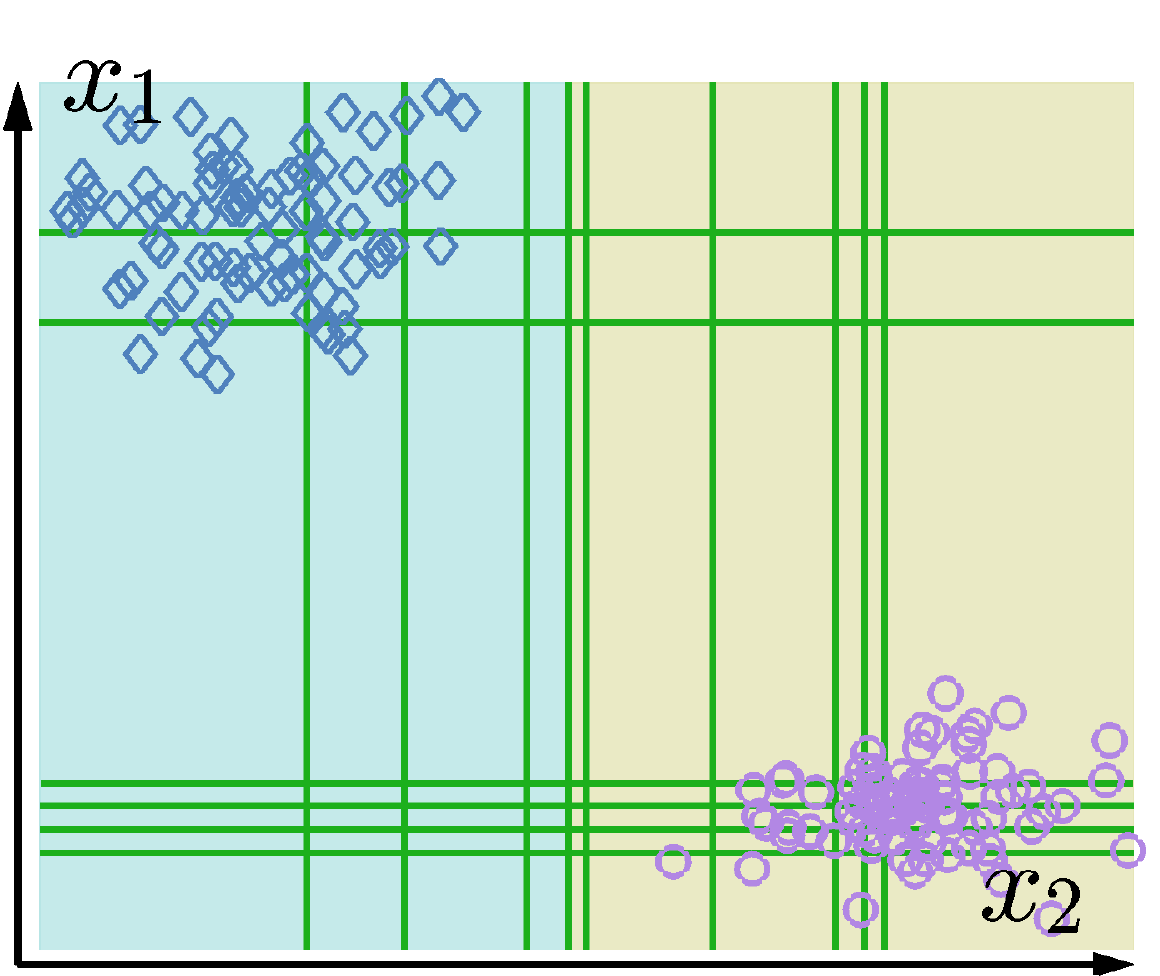}
	}
   \caption{
   An illustration of clustering toy data
   with a clustering forest. 
   (a) Original toy data are labelled as class $1$, 
   whilst (b) the pseudo-points (red $+$)
	as class $2$. 
   (c) A clustering forest performs two-class classification in the augmented space. 
   (d) The final data partitions on the original data. }
\label{fig:view_RF}
\end{figure*}

\noindent \textbf{Classification forests} - 
A general form of random forests is the classification forests.
A classification forest~\cite{BreimanML01,Schulter_2013_CVPR} 
is an ensemble of $T_\mathrm{class}$ binary decision trees 
$\mathcal{T}(\mathbf{x})$: $\mathcal{X} \rightarrow \mathbb{R}^K$,
with $\mathcal{X}$ the $d$-dimensional feature space, and
$\mathbb{R}^K = [0, 1]^K$ denoting the space of class probability distribution
over the label space $\mathcal{L} = \{1,\dots,K\}$.

Decision trees are learned independently of each other, 
each with a random subset ${X}_t$ of the training samples ${X} = \{\mathbf{x}_i\}$, i.e. bagging~\cite{BreimanML01}.
Growing a decision tree involves a recursive node splitting procedure 
until some stopping criterion is satisfied, 
e.g. leaf nodes are formed when no further split can be achieved given the objective function, 
or the number of training samples arriving at a node is smaller than the predefined node size, $\phi$.
Small $\phi$ leads to deep trees. 
We set $\phi=2$ in our experiments for capturing sufficiently
fine-grained data structure.
At each leaf node, the class probability distribution 
is then estimated based on the labels of the arrival samples.

The training of each internal/split node is a process of binary split function optimisation, defined as
\begin{equation}
	\mathrm{h}(\mathbf{x}, \boldsymbol{\vartheta}) = 
    \left\{
	\begin{array}{l l}
    	0 	& \quad \text{if ${x}_{\vartheta_1} < \vartheta_2$,} \\
		1 	& \quad \text{otherwise.} \\
	\end{array} \right.
\label{eqn:split_function}
\end{equation}
This split function is parameterised by two parameters $\boldsymbol{\vartheta} = \left[\vartheta_1, \vartheta_2\right]$: 
(i) a feature dimension ${x}_{\vartheta_1}$ with $\vartheta_1 \in \{1,\dots, d\}$, and 
(ii) a feature threshold $\vartheta_2 \in \mathbb{R}$. 
All samples of a split node $s$ will be channelled
to either the left $l$ or right $r$ child nodes, 
according to the output of Eqn.~(\ref{eqn:split_function}).

The optimal split parameter $\boldsymbol{\vartheta}^*$ is chosen via
\begin{equation}
\boldsymbol{\vartheta}^* = \argmax_{\Theta} \Delta \mathcal{I}_\mathrm{class},
\label{eqn:split_parameter_optimisation}
\end{equation}
where $\Theta = {\left\{ \boldsymbol{\vartheta}^i \right\}_{i=1}^{m_\mathrm{try}(|S|-1)}}$ represents a parameter set
over $m_\mathrm{try}$ randomly selected features, 
with $S$  the sample set reaching the node $s$.
The cardinality of a set is given by $|\cdot|$.
Typically, a greedy search strategy is exploited to identify $\boldsymbol{\vartheta}^*$.
The information gain $\Delta \mathcal{I}_\mathrm{class}$ is formulated as
\begin{equation}
\Delta \mathcal{I}_\mathrm{class} = \mathcal{I}_s - 
\frac{|L|}{|S|} \mathcal{I}_l - \frac{|R|}{|S|} \mathcal{I}_r,
\label{eqn:info_gain}
\end{equation}
where $L$ and $R$ denote the sets of data routed into $l$ and $r$, 
and $L \cup R = S$. 
The information gain $\mathcal{I}$ can be computed as either the entropy or 
Gini impurity~\cite{breiman1984classification}.

\vspace{0.1cm}
\noindent \textbf{Clustering forests} -  
In contrast to classification forests, 
clustering forests require no ground truth label information during the training phase.
A clustering forest consists of $T_\mathrm{clust}$ binary decision trees. 
The leaf nodes in each tree define a spatial partitioning of the training data.
Interestingly, the training of a clustering forest can be performed
using the classification forest optimisation approach 
by adopting the pseudo two-class algorithm~\cite{BreimanML01,LiuCIKM2000,shi2006unsupervised}.
Specifically, we add $N$ pseudo samples 
$\mathbf{\bar{x}} = \{ \bar{x}_1, \dots, \bar{x}_\mathit{d} \}$ 
(Fig.~\ref{fig:view_RF}(b)) into the original data space ${X}$ (Fig.~\ref{fig:view_RF}(a)),
with $\bar{x}_i \sim \mathrm{Dist}(x_i)$ sampled from certain distributions
$\mathrm{Dist}(x_i)$. 
In the proposed model, we adopt the empirical marginal distributions 
of the feature variables owing to its favourable performance~\cite{shi2006unsupervised}. 
With this data augmentation strategy, 
the clustering problem becomes a canonical classification problem 
that can be solved by the classification forest training method as discussed above. 
The key idea behind this algorithm is 
to partition the augmented data space into dense and sparse regions (Fig.~\ref{fig:view_RF}(c-d))~\cite{LiuCIKM2000}.

\subsection{Multi-Source Clustering Forest}
\label{sec:method@CC_forest}

Conventional clustering forests assumes only homogeneous data sources such as pure imagery-based features. In contrast, the proposed Multi-Source Clustering Forest can take heterogeneous sources as input.
In particular, the proposed model uses visual features as splitting variables to grow Multi-Source Clustering trees (MSC-trees) as in Eqn.~(\ref{eqn:split_function}), and exploits non-visual information as additional data to help determining the $\boldsymbol{\vartheta} = \left[\vartheta_1, \vartheta_2\right]$. In this way, auxiliary non-visual information is used, in addition to visual data, to guide the tree formation.

Formally, we define a new joint information gain function for node splitting during training MSC-trees as:
\begin{equation}
	\Delta \mathcal{I} = 
	\underbrace{\alpha_v \frac{\Delta\mathcal{I}_v}{\mathcal{I}_{v0}}}_{\text{visual}} + 
	\underbrace{\sum_{j = 1}^{\mathit{m}}\alpha_{j} \frac{\Delta \mathcal{I}_{j}}{\mathcal{I}_{j0}}}_{\text{non-visual}} + 
	\underbrace{\alpha_t \frac{\Delta \mathcal{I}_{t}}{\mathcal{I}_{t0}}}_{\text{temporal}}.
	\label{eqn:info_gain_our}
\end{equation}
Similar to Eqn.~(\ref{eqn:info_gain}), the optimal parameter corresponds to 
the split with the maximal $\Delta \mathcal{I}$. 
This formulation defines the best data split across the joint space of multi-source data,
beyond visual domain alone.
All the terms in Eqn.~(\ref{eqn:info_gain_our}) are interpreted as below.

\vspace{0.1cm}
\noindent \textit{Visual term}: 
$\Delta\mathcal{I}_v = \Delta \mathcal{I}_\mathrm{class}$ (Eqn.~(\ref{eqn:info_gain})) 
denotes the information gain in visual domain.
Precisely, this measure is computed from the pseudo class labels.
Therefore, it reflects the visual data structure characteristics given that
the pseudo data samples are drawn from the marginal feature distributions (Section~\ref{sec:method@RF_background}).
In this study we utilise the Gini impurity $\mathcal{G}$~\cite{breiman1984classification}
to estimate $\Delta \mathcal{I}_\mathrm{class}$ by setting $\mathcal{I} = \mathcal{G}$ in Eqn.~(\ref{eqn:info_gain}) due to its simplicity and efficiency.
The Gini impurity is computed as 
	$\mathcal{G} = \sum_{i \neq j} p_i p_j$,
with $p_i$ and $p_j$ being the proportion of samples 
belonging to the $i$th and $j$th category in a split node $s$.
High value in $\mathcal{G}$ indicates pure category distribution.

\vspace{0.1cm}
\noindent \textit{Non-visual term}: 
This is a new term we introduce as auxiliary information on visual term.
%
More specifically, $\Delta \mathcal{I}_{j}$ denotes the information gain 
in the $j$th non-visual data.
A non-visual source can be either categorical or continuous.
For a categorical non-visual source, similar to visual term 
we use the Gini impurity $\mathcal{G}$ as its data split measure criterion.
In the case of non-visual source with continuous values,  
we adopt least squares regression~\cite{breiman1984classification} to enforce continuity in the clustering space:
\begin{equation}
\mathcal{R} = \frac{1}{|S|} \sum_{i=1}^{|S|} (y_{i,j} - \frac{1}{|S|}\sum_{i=1}^{|S|}y_{i,j})^2,
\end{equation}
where $y_{i,j}$ represents the value in the $j$th non-visual space 
associated with the $i$th sample $\mathbf{x}_i \in S$, and $S$ is 
the set of samples reaching node $s$.
That is $\Delta \mathcal{I}_{j} = \mathcal{R}$. 


\vspace{0.1cm}
\noindent \textit{Temporal term}:
We also add a temporal smoothness gain $\Delta \mathcal{I}_{t}$ 
to encourage temporally adjacent video clips to be grouped together.
This temporal information helps in mining visual data structure.

The information gain by different sources may live in very disparate ranges 
due to the different natures of source,
each term of Eqn.~(\ref{eqn:info_gain_our}) is therefore normalised 
by its initial data impurity
denoted by $\mathcal{I}_{v0}$, $\mathcal{I}_{j0}$, 
and $\mathcal{I}_{t0}$.
These impurities are obtained at the root node of every MSC-tree.
The source weights are denoted by $\alpha_{v}$, $\alpha_{i}$, and $\alpha_t$ accordingly,
holding $\alpha_v + \sum_{i = 1}^{\mathit{m}}\alpha_{i} + \alpha_t = 1$.
We set $\alpha_{v}=0.5$ obtained by cross-validation. A detailed analysis on $\alpha_{v}$ is given in Section~\ref{sec:Exp@tagging}.
For non-visual and temporal information, we uniformly assign
$\alpha_t = \alpha_{i} = \frac{1-\alpha_v}{m+1}$ since 
their importance is not known \textit{in prior}, with $m$ the number of non-visual sources.

\vspace{0.1cm}
\noindent \textbf{The role of different source data} - 
Given the main role and much more stable provision of the visual source in video understanding, non-visual data are regarded as auxiliary information over visual source. During the training of MSC-Forest, 
the split functions (Eqn.~(\ref{eqn:split_function})) are defined on visual features, but $\boldsymbol{\vartheta} = \left[\vartheta_1, \vartheta_2\right]$ is collectively determined by visual features and the associated 
non-visual as well as temporal information (i.e. the non-visual and temporal term in Eqn.~(\ref{eqn:info_gain_our})).
Alternatively, one can think of that the \textit{main} visual data source is `completely-visible'
to the MSC-Forest since it is needed during both forest training and evaluation,
whilst the \textit{auxiliary} non-visual data are `half-visible' in that 
they are exploited as side information for embedding their knowledge into the MSC-tree growing during model training but not required any more during the MSC-Forest evaluation (due to their restricted availability as explained in Section~\ref{sec:introduction}).

\vspace{0.1cm}
\noindent \textbf{Joint information gain} - 
We interpret the intrinsic advantage of the joint information gain defined by Eqn.~(\ref{eqn:info_gain_our}), 
with comparison against the na\"{i}ve feature concatenation strategy.
With the latter scheme, the information gain (Eqn.~(\ref{eqn:info_gain}))
is directly estimated in a heterogeneous joint space 
where visual, non-visual and temporal data are mixed together.
This would suffer from the heteroscedasticity problem, as discussed in Section~\ref{sec:introduction}.
Instead, Eqn.~(\ref{eqn:info_gain_our}) overcomes this challenge
by modelling different sources via separate information gain terms, 
resulting in a more balanced exploitation of multi-source data.
In this way, the proposed joint information gain of multi-source data
encourages more appropriate visual data separation both visually and semantically.
This formulation is the essential contribution of our proposed MSC-Forest model.

\vspace{0.1cm}
\noindent \textbf{The merits of MSC-Forest} - 
The formulation in Eqn.~(\ref{eqn:info_gain_our}) brings two unique benefits:
(A) Thanks to the information gain optimisation, the influences of visual and non-visual domains on data partitioning can be better balanced compared to na\"{i}ve feature concatenation.
(B) Eqn.~(\ref{eqn:split_parameter_optimisation}) and Eqn.~(\ref{eqn:info_gain_our}) together provide a mechanism to discover strongly correlated heterogeneous source pairs and to exploit joint information gain of such correlated pairs for data partitioning. In other words, 
only selective visual features 
(Eqn.~(\ref{eqn:split_parameter_optimisation})) that yield high information gain collectively with non-visual information (Eqn.~(\ref{eqn:info_gain_our})) will contribute to the MSC-tree growing. Such a mechanism cannot be realised using the conventional clustering forests~\cite{BreimanML01,LiuCIKM2000}.
We shall demonstrate the multi-source correlation discovered by our proposed MSC-Forest in experiments (Section~\ref{sec:model_visualisation}).

\subsubsection{Coping with Partial/Missing Non-Visual Data}
\label{sec:method@adaptive_weighting}
We introduce a new adaptive weighting mechanism 
to dynamically deal with the inevitable partial/missing non-visual data\footnote{
There exist missing data filling algorithms utilised in conventional random forests,
e.g. for the missing value of one feature in one class, the median value (continuous) or the most frequent category (discrete) of this feature over the current class can be used as the estimation~\cite{breiman2003rf}. 
Whilst a similar strategy is possible to apply on our MSC-Forest, 
we consider an alternative 
by proposing an effective adaptive weighting algorithm
in order not to further introduce noisy training data.
}.
%
Specifically, 
%
when some non-visual data are missing and 
suppose the missing proportion of the $i$th non-visual type 
in the training set $\mathbf{X}_t$ for MSC-tree $t$ is $\delta_i$,
we reduce its weight from $\alpha_i$ to $\alpha_i - \delta_i\alpha_i$. 
The total reduced weight $\sum_i \delta_i\alpha_i$ is then distributed evenly 
to the weights of all sources to ensure 
$\alpha_v + \sum_{i = 1}^{\mathit{m}}\alpha_{i} + \alpha_t = 1$. 
This linear adaptive weighting method produces satisfactory results in our experiments. 
%

%

\subsubsection{Model Complexity}
\label{sec:method@complexity}


The upper-bound learning complexity of a whole MSC-Forest can be examined from its constituent parts, i.e. at tree- and node-levels.
Formally, given a MSC-tree $t$,
we denote the set of all the split nodes as $\Pi_t$ and 
the sample subset used for training a split node $j \in \Pi_t$ as $S_j$. 
The training complexity of $j$-th node is given by
$m_\mathrm{try} (|S_j|-1) u$,
when a greedy search algorithm is adopted, 
with $m_\mathrm{try}$ the number of features attempted to partition $S_j$,
and $u$ the running time of conducting one data splitting operation.
Consequently, the overall computational cost of learning a MSC-Forest 
can be computed as
\begin{equation}
\Omega = \sum^{T_\mathrm{clust}}_t
\sum_{j \in \Pi_t} m_\mathrm{try} (|S_j| - 1) u = 
m_\mathrm{try} u \sum^{T_\mathrm{clust}}_t \sum_{j \in \Pi_t} (|S_j| - 1) .
\label{eqn:complexity_learning_forest}
\end{equation}
%
The value of parameter $m_\mathrm{try}$ is identical across all MSC-trees. 
The learning time is thus determined by 
(1) the value of $u$, and
(2) the factor that we name as \textit{tree fan-in}
\begin{equation}
\Phi(t) = \sum_{j \in \Pi_t} |S_j-1|.
\end{equation}
Clearly, $u$ of a MSC-Forest is larger than that of conventional forests 
since we need to compute additional information gains of non-visual and temporal information (Eqn.~(\ref{eqn:info_gain_our})).
On the other hand, the value of $\Phi(t)$ primarily relies on 
the tree structure/topological characteristics~\cite{martin1997exact}:
a balanced and shallower tree has smaller $\Phi(t)$,
thus the tree shall be more efficient in training and inference on previously-unseen samples, 
in that the paths from the root to leaf nodes
are relatively shorter.
In Section~\ref{sec:exp@complexity}, we will show that the additional non-visual information encourages more balanced and shallower decision trees than learning from single visual source alone.

\subsection{Latent Multi-Source Data Structure Discovery}
\label{sec:method@latent_data_structure_discovery}
The multi-source feature space has high dimension (over $2000$ dimensions). 
This makes learning data structure by clustering computationally difficult.
To this end, we consider spectral clustering on manifold 
to discover latent clusters in a lower dimensional space. 
Fig.~\ref{fig:data_structure_discovery} depicts the pipeline 
of our video data clustering approach based on the learned MSC-Forest.

The spectral clustering~\cite{ZelnikNIPS2004} groups data 
using eigenvectors of an affinity matrix derived from the data.
The goodness of the resulting cluster formation primarily relies on the quality of the input affinity matrix
which reflects and embeds the essential data structures~\cite{zhu2014constructing}.
Below we describe the details of constructing multi-source referenced affinity matrix 
from MSC-Forest. 
Intuitively, the multi-source learning nature of MSC-Forest renders 
its data similarity measure sensitive to the joint knowledge from diverse source data.

The learned MSC-Forest offers an effective way to derive the required affinity matrix.
Specifically, each individual tree within the MSC-Forest 
partitions the training samples at its leaves 
$\ell(\mathbf{x})$: $\mathbb{R}^d \rightarrow \mathbf{L} \subset \mathcal{N}$, 
where $\ell$ represents a leaf index and 
$\mathbf{L}$ refers to the set of all leaves in a given tree. 
For each MSC-tree, we first compute a tree-level 
$N \times N$ affinity matrix $A^t$ with elements defined as 
$A_{i,j}^t = \exp^{-\mathrm{dist} (\mathbf{x}_i, \mathbf{x}_j)}$ where
\begin{equation}
	\mathrm{dist}(\mathbf{x}_i, \mathbf{x}_j) = \left\{
		\begin{array}{l l}
    		0 		& \quad \text{if $\ell(\mathbf{x}_i) = \ell(\mathbf{x}_j)$},	\\
    		+\infty & \quad \text{otherwise}.	\\
		\end{array} \right.
		\label{eqn:pair_dist}
\end{equation}
We assign the maximum affinity (affinity=1, distance=0) between points $\mathbf{x}_i$ 
and $\mathbf{x}_j$ if they fall into the same leaf, 
and the minimum affinity (affinity=0, distance=1) otherwise. 
A smooth affinity matrix can be obtained through
averaging all the tree-level affinity matrices
\begin{equation}
A = \frac{1}{T_\mathrm{clust}}\sum_{t=1}^{T_\mathrm{clust}} A^t,
\label{eqn:affinity_forest_our}
\end{equation}
Eqn.~(\ref{eqn:affinity_forest_our}) is adopted as the ensemble model of MSC-Forest due to
its advantage of suppressing the noisy tree predictions, though other alternatives such as
the product of tree-level predictions are possible~\cite{criminisi2012decision}.
We then construct a sparse $k$-NN graph, 
whose edge weights are defined by $A$ (Fig.~\ref{fig:data_structure_discovery}(c)).

Subsequently, we symmetrically normalise $A$ to obtain $\mathcal{S} = \mathit{D}^{-\frac{1}{2}} A \mathit{D}^{-\frac{1}{2}}$, where $ \mathit{D} $ denotes a diagonal degree matrix with elements 
${D}_{i,i} = \sum_j^{N}{A_{i,j}}$.
Given $\mathcal{S}$, we perform spectral clustering to 
discover the latent clusters of training clips 
with the number of clusters automatically determined through 
analysing the eigenvector structure~\cite{ZelnikNIPS2004}.
Each training clip $\mathbf{x}_i$ is then 
assigned to a cluster $c_i \in \mathcal{C}$,
with $\mathcal{C}$ the set of all clusters.

The learned clusters group similar clips both visually and semantically, 
with each of the clusters associated with 
a unique distribution for each non-visual data (Fig.~\ref{fig:data_structure_discovery}(d)). 
We denote the distribution of the $i$th non-visual data type of the cluster $c$ as
\begin{equation}
p(y_i|c) \propto \sum\nolimits_{\mathbf{x}_j \in \mathbf{X}_c} p(y_i|\mathbf{x}_j),
\label{eqn:nonvisual_data_pdf}
\end{equation} 
where ${\mathbf{X}_c}$ represents the set of training samples in $c$.
These multi-source data clusters form a component of our multi-source model (Fig.~\ref{fig:overview}).

\section{Semantic Video Summarisation}
\label{sec:method@summarisation}

In Section~\ref{sec:method} we presented multi-source data clustering by learning a Multi-Source Clustering Forest (MSC-Forest), resulting in a semantic cluster formation. 
Once this multi-source model is learned, it can be deployed for semantic video summarisation.
Specifically, we follow the established approach of summarising videos by clustering~\cite{truong2007video} but with the introduction of two noticeable differences in our method.

First, our \textit{video summary is multi-source referenced}. 
%
Specifically, the MSC-Forest
is trained on heterogeneous sources, its optimised split functions $\{\mathrm{h}\}$ (Eqn.~(\ref{eqn:split_function})) therefore implicitly capture the complex multi-source structures.
%
When one deploys the trained model for
content summarisation of previously-unseen video data, 
the model only needs to take visual inputs without
any non-visual data sources. And yet it is able to induce video content
partitions that not only correspond to visual feature similarities,
but also are consistent with meaningful non-visual semantic interpretations.
Second, our \textit{video summary is automatically tagged} as the
result of model inference. 
This is made possible through exploiting the non-visual
data distributions associated with the discovered clusters on the training data (see Eqn.~(\ref{eqn:nonvisual_data_pdf}) and Fig.~\ref{fig:data_structure_discovery}(d)). 
%
Below we discuss the details of generating a semantic video summary.

\begin{figure*}[t]
\centering
\includegraphics[width=0.99\linewidth]{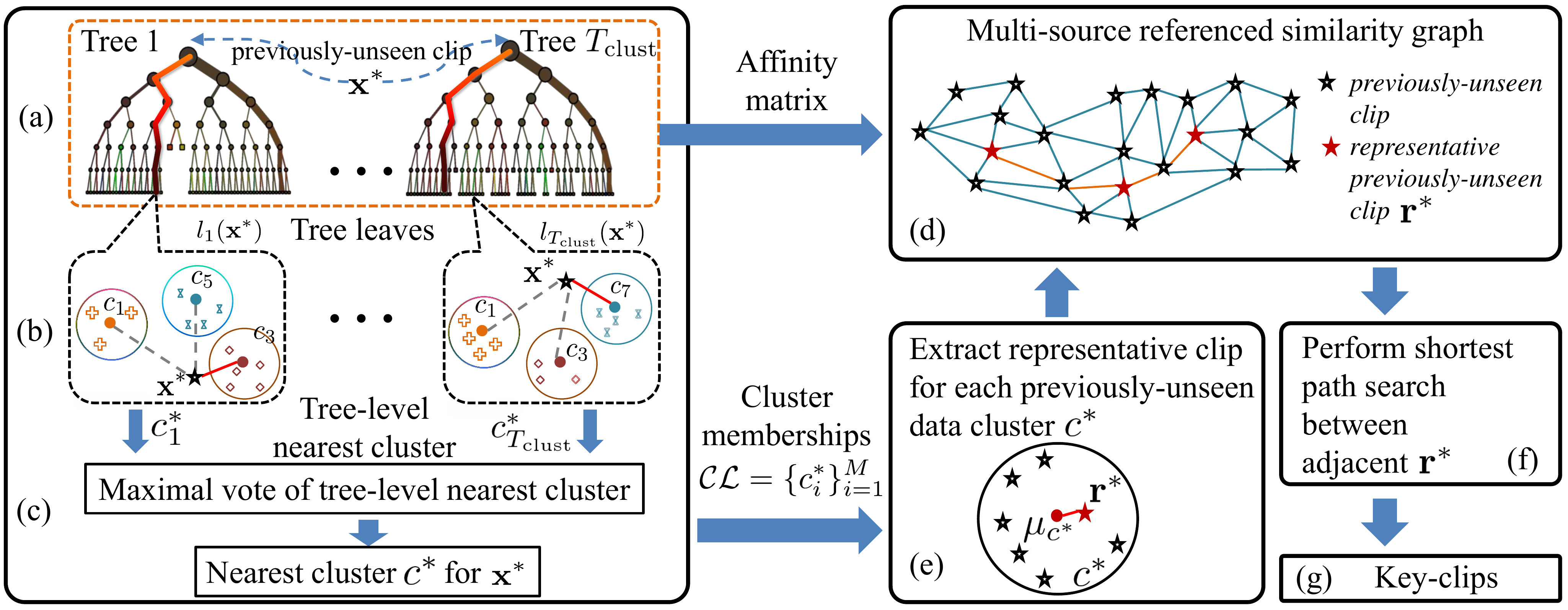}
\caption{\footnotesize The pipeline of our multi-source referenced 
    key-clips detection algorithm. 
(a) Channel a clip $\mathbf{x}^*$ into MSC-trees. 
(b) Search tree-level nearest clusters of $\mathbf{x}^*$, 
hollow circle denotes cluster. 
(c) Predict the final nearest cluster.
A red $\star$ depicts a representative previously-unseen clip.
}
\label{fig:keyclips_extraction}
\end{figure*}

\subsection{Key-Clip Extraction and Composition}
\label{sec:method@keyVisualContentExtraction}

Suppose we are given a previously-unseen surveillance video footage without meta-data tagging/script. The video is pre-processed by
segmenting it into a set of $M$ either overlapping or non-overlapping
short clips $\{\mathbf{x}_i^*\}_{i=1}^{M}$ with equal duration.
Our aim is to first assign cluster membership to each previously-unseen clip using the trained multi-source model, and then select key-clips from the resulting clusters\footnote{
It is worth noticing that the purpose of this clustering step is completely different from 
the multi-source data clustering during model training, as presented in Section~\ref{sec:method@latent_data_structure_discovery}.
The latter is a component of our multi-source model training pipeline (Fig.~\ref{fig:data_structure_discovery}), 
whilst the former aims at revealing the latent structure over testing data for video summarisation.
}.
The chosen key-clips are then chronologically ordered to construct a video summary.

%

\vspace{0.1cm}
\noindent 
\textbf{Clustering previously-unseen video clips} - 
Inferring cluster memberships of previously-unseen clips is an intricate task.
A straightforward method is to assign cluster membership by identifying the nearest cluster $c^* \in \mathcal{C}$ to a sample $\mathbf{x}^*$, where $\mathcal{C}$ represents the set of clusters we discovered in Section~\ref{sec:method@latent_data_structure_discovery}.
However, we found this hard cluster assignment strategy
susceptible to outliers in $\mathcal{C}$ and source noise.
To mitigate this problem, we consider an alternative approach by utilising
the MSC-Forest tree structures for soft cluster assignment. 
This is more robust to either
source noise or outliers.

Fig.~\ref{fig:keyclips_extraction} depicts the soft cluster assignment pipeline.
First, we trace the leaf $\ell_t(\mathbf{x}^*)$ of each tree $t$  
where $\mathbf{x}^*$ falls by
channelling $\mathbf{x}^*$ into the tree (Fig.~\ref{fig:keyclips_extraction}(a)).
This step is critical as it establishes a connection for $\mathbf{x}^*$ with an appropriate training subset $\mathbf{X}_{\ell_t(\mathbf{x}^*)}$ using the split functions $\{\mathrm{h}\}_t$ optimised by multi-source data.
Here, $\mathbf{X}_{\ell_t(\mathbf{x}^*)}$ represents the set of training samples associated with $\ell_t(\mathbf{x}^*)$. The set is consistent with $\mathbf{x}^*$ both visually and semantically since they encompass identical response w.r.t $\{\mathrm{h}\}_t$.

Second, we retrieve the cluster membership $C_t = \left\{ c_i \right\} \subset \mathcal{C}$ of $\mathbf{X}_{\ell_t(\mathbf{x}^*)}$,
against which we search for the
tree-level nearest cluster $c^*_t$ for $\mathbf{x}^*$ (Fig.~\ref{fig:keyclips_extraction}(b)) via
\begin{equation}
	c^*_t = \argmin\nolimits_{c \in C_t} || {\mathbf{x}}^* - \mu_{c}||,
	\label{eqn:NN_cluster}
\end{equation}
with $t$ the tree index, and $\mu_{c}$ the centroid of cluster $c$, estimated as 
\begin{equation}
	\mu_{c} = \frac{1}{|\mathbf{X}_c|} \sum_{\mathbf{x}_i \in {\mathbf{X}_c}} \mathbf{x}_i,
	\label{eqn:clust_center}
\end{equation}
where ${\mathbf{X}_c}$ represents the set of training samples in $c$.
Performing nearest cluster search within $C_t$ rather than 
the whole cluster space $\mathcal{C}$ brings a key benefit:
since the search space is constrained by MSC-tree, it is more meaningful and also less noisy than the entire space $\mathcal{C}$, leading to more accurate $c^*_t$ estimation.

Once we obtain all tree-level nearest clusters from all the trees in the forest, $\left\{c^*_t \right\}_{t=1}^{T_\mathrm{clust}}$ , the final nearest cluster $c^*$ is obtained as the one with maximal votes from all the trees (Fig.~\ref{fig:keyclips_extraction}(c)) 
\begin{equation}
c^* = \max{\left\{c^*_t \right\}_{t=1}^{T_\mathrm{clust}}} 
\end{equation}
By repeating the above steps on all previously-unseen clips $\{\mathbf{x}_i^*\}_{i=1}^{M}$, we obtain their cluster labels as $\mathcal{CL} = \{ c_i^*\}_{i=1}^{M}$ (Fig.~\ref{fig:keyclips_extraction}(e)).


\vspace{0.1cm}
\noindent
\textbf{Extracting key-clips} -  
With the assigned cluster memberships $\mathcal{CL}$ on all previously-unseen clips, 
the key-clip of a previously-unseen video data cluster $c^*$ can be represented by the representative previously-unseen clip $\mathbf{r}^*$ 
that is closest to the cluster centroid $\mu_{c^*}$ (Fig.~\ref{fig:keyclips_extraction}(e)).
Concatenating these key-clips chronologically establishes a visual summary.
Such a summary, however, is likely to be discontinuous in preserving
visual context therefore non-smooth visually due to abrupt changes between adjacent key-clips.
To enforce some degrees of smoothness in the visualisation of video
summary whilst minimising redundancy, we adopt a shortest path
strategy~\cite{boccaletti2006complex} 
to induce an optimal path 
between two temporally-adjacent representative $\mathbf{r}^*$
on a graph $G$. 
This approach produces a visually more coherent video summary whilst discards as much redundancy as possible.

More precisely, we construct a graph $G =(V, E)$, where $V$ and $E$ indicate the set of previously-unseen video clip vertices and edges (Fig.~\ref{fig:keyclips_extraction}(d)).
The weights of edges can be efficiently estimated using Eqn.~(\ref{eqn:pair_dist}) and~(\ref{eqn:affinity_forest_our}).
Note that the graph $G$ is also multi-source referenced since it is derived from our multi-source MSC-Forest model.
We then perform shortest path search between 
temporally-adjacent $\mathbf{r}^*$ on $G$ (Fig.~\ref{fig:keyclips_extraction}(f)) and  
all the samples that lie on the shortest paths compose 
the final key-clip set $\mathcal{K}$ (Fig.~\ref{fig:keyclips_extraction}(g)).

\subsection{Video Tagging}
\label{sec:method@tagging}

\begin{algorithm}
	\caption{ Infer non-visual tags of previously-unseen clips.}
	\label{alg:tagging_alg}
	{
		\SetAlgoLined
		\KwIn{A previously-unseen clip $\mathbf{x}^*$, a trained MSC-Forest, training data clusters $\mathcal{C}$;}
		\KwOut{Predicted tag $\hat{y}_i$;}

		\textbf{Initialisation:} \\
		\quad Compute $p(y_i|c)$ for each training data cluster (Eqn.~(\ref{eqn:nonvisual_data_pdf})); \\
		\quad Compute cluster centroid $\mu_{c}$ (Eqn.~(\ref{eqn:clust_center})); \\
	
 		\textbf{Non-Visual Tag Inference:} \\
		\For{$t\leftarrow 1$ \KwTo $T_{\mathrm{clust}}$}
		{
  			Trace the leaf $\ell_t(\mathbf{x}^*)$ where $\mathbf{x}^*$ falls (Fig.~\ref{fig:keyclips_extraction}(a)); \\
  			Retrieve the training samples $\mathbf{X}_{\ell_t(\mathbf{x}^*)}$ associated with $\ell_t(\mathbf{x}^*)$; \\
  			Obtain the clusters $C_t = \{c_i\} \subset \mathcal{C}$ of $\mathbf{X}_{\ell_t(\mathbf{x}^*)}$; \\
  			Search the tree-level nearest cluster $c_t^*$ of $\mathbf{x}^*$ within $C_t$ (Eqn.~(\ref{eqn:NN_cluster})); \\
 		}
 		
 		Estimate tag distribution $p(y_i|\mathbf{x}^*)$ (Eqn.~(\ref{eqn:tag_dist_by_forest})); \\
 		Compute the final tag $\hat{y_i}$ (Eqn.~(\ref{eqn:final_tag})).
	}
\end{algorithm}

Summarising video with high-level interpretation requires plausible semantic content inference from video data $\mathbf{x}^*$. 
We derive a tree-structure aware tag inference algorithm capable of predicting tag types 
same as training non-visual data,
based on the learned MSC-Forest and discovered training data clusters.
%
Specifically, we first obtain the tree-level nearest cluster $c^*_t$ 
of a previously-unseen sample $\mathbf{x}^*$ using Eqn.~(\ref{eqn:NN_cluster}).
Second, the $p(y_i|c^*_t)$ associated with $c^*_t$ is utilised 
as the tree-level non-visual tag estimation for the $i$th non-visual data type.
To achieve a smooth prediction, we average all $p(y_i|c=c^*_t)$
obtained from individual trees as
\begin{equation}
	p(y_i|\mathbf{x}^*) = \frac{1}{T_\mathrm{clust}} \sum\nolimits_{t=1}^{T_\mathrm{clust}}p(y_i|c^*_t).
	\label{eqn:tag_dist_by_forest}
\end{equation}
The final tag $\hat{y}_i$ for the $i$th non-visual type is obtained as
\begin{equation}
	\hat{y}_i = \argmax\nolimits_{y_i} p(y_i|\mathbf{x}^*).
	\label{eqn:final_tag}
\end{equation}
With the above steps, we can estimate all $m$ non-visual tags $\hat{y}_i$s 
with $i \in \{1,\dots, m$\}. 
The procedure of our 
tagging algorithm is summarised 
in Algorithm~\ref{alg:tagging_alg}.

Given the extracted key-clips $\mathcal{K}$
  and automatic assignment of
non-visual semantic tags (Eqn.~(\ref{eqn:final_tag})),
we can now construct a
video summary by chronologically concatenating each clip
$\mathbf{x}^* \in \mathcal{K}$ with smooth inter-clip transition, \eg~crossfading, 
and labelling each clip with their inferred semantic tags. 

\section{Experimental Settings}
\label{sec:Exp_Settings}
\begin{figure*}[ht] 
\centering
	\includegraphics[width=0.7\linewidth]{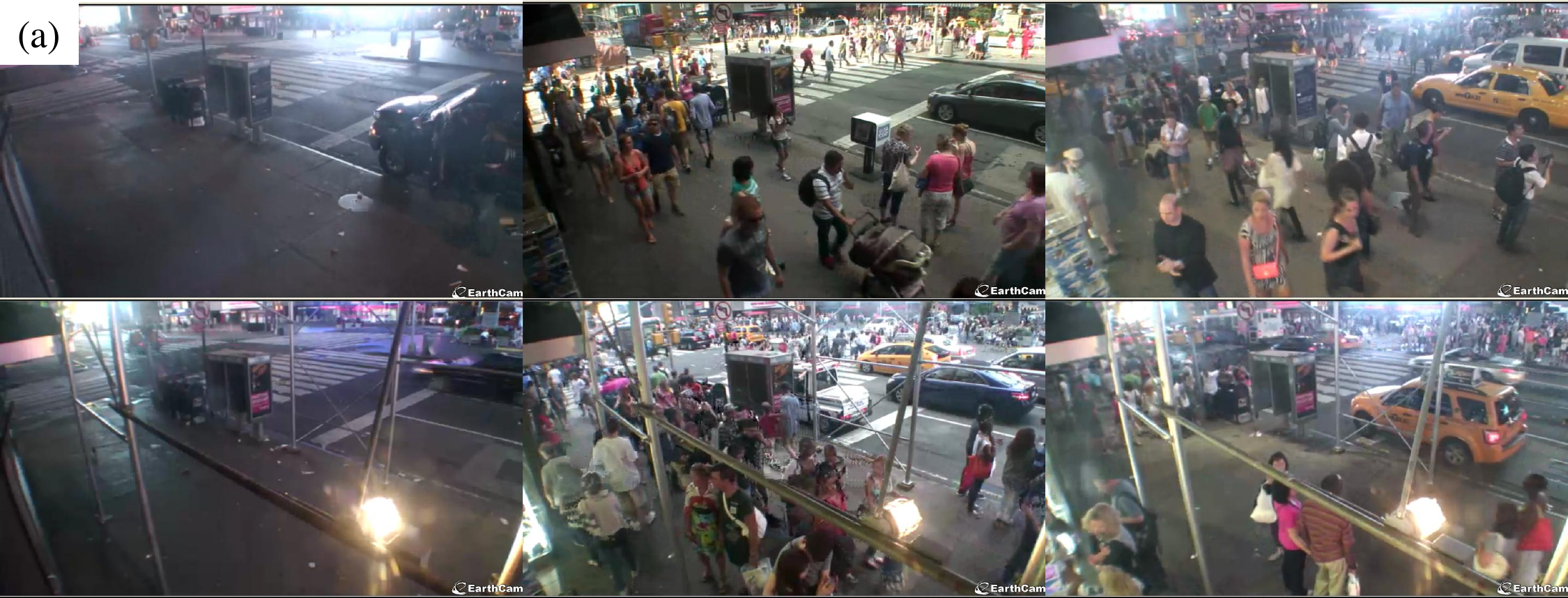} 
	\includegraphics[width=0.7\linewidth]{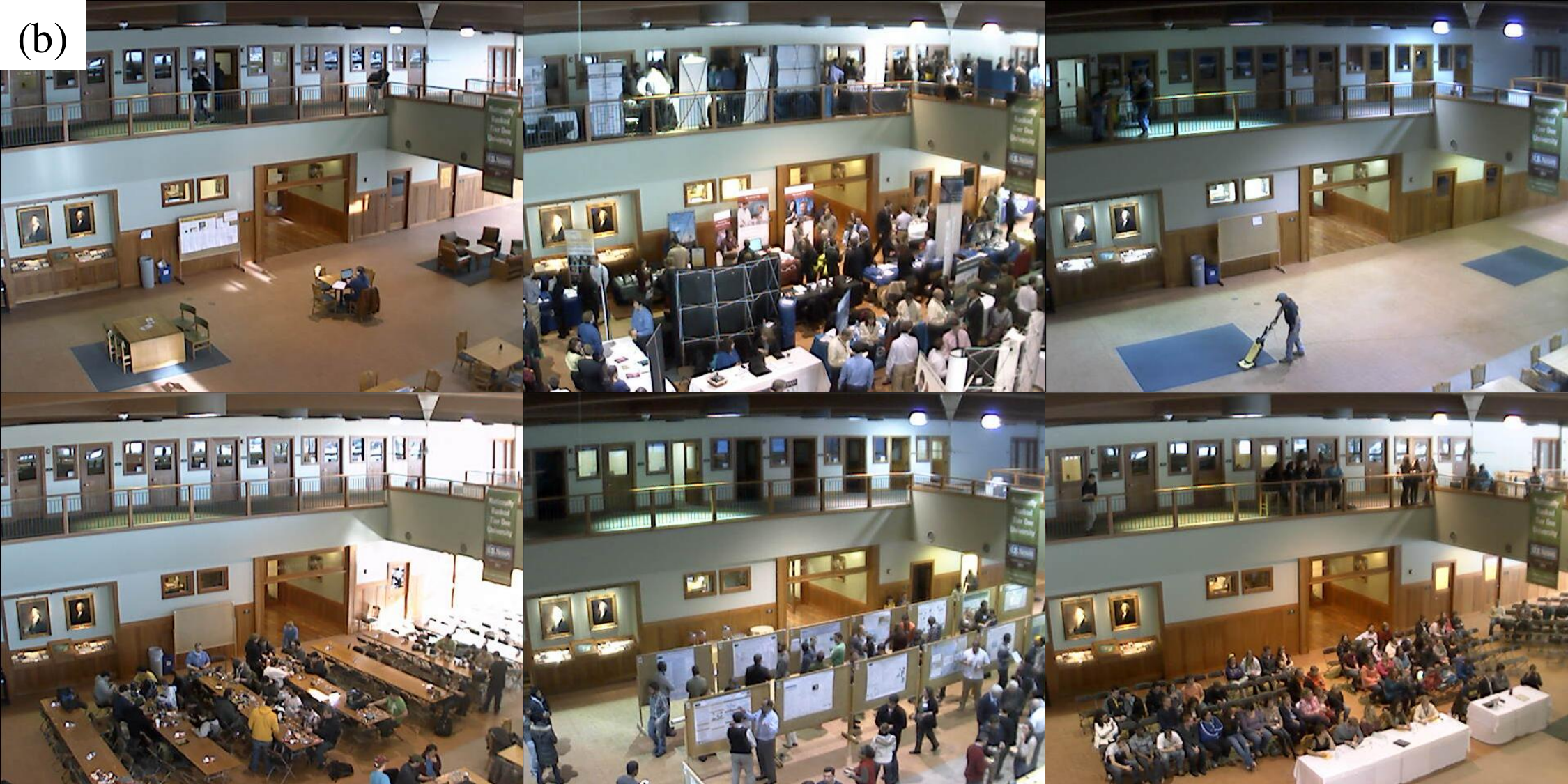}
\caption{\footnotesize Examples of the (a) TISI and (b) ERCe datasets.}
\label{fig:dataset_eg}
\end{figure*}

\noindent \textbf{Datasets} - 
We conducted experiments on two datasets
collected from publicly
accessible webcams that feature an outdoor and an indoor scene
respectively: (1) the TImes Square Intersection (TISI) dataset, 
and (2) the Educational Resource Centre (ERCe) dataset\footnote{
Datasets available: www.eecs.qmul.ac.uk/\%7Exz303/download.html}. 
%
%
There are a total of $7324$ video clips spanning over $14$ days in the TISI dataset, whilst a total of 13817 clips were collected across a period of two months in the ERCe dataset.
Each clip has a duration of $20$ seconds.
The details of the datasets and training/deployment partitions are given in Table~\ref{table:dataset_details}. Example frames are shown in Fig.~\ref{fig:dataset_eg}.

The TISI dataset is challenging due to severe inter-object occlusion, complex behaviour patterns, and large illumination variations caused by both natural and artificial light sources at different day time. 
The ERCe dataset is non-trivial due to a wide range of physical events
involved that are characterised by large changes in environmental
setup, participants, crowdedness, and intricate activity
patterns.

\begin{table} [h] \footnotesize
\caption{ Details of datasets. FPS = frames per second.}
\label{table:dataset_details}
\centering
\begin{tabular}{c|c|c|c|c}
 - & Resolution & FPS & \# Training Clip & \# Deployment Clip \\
\hline
\hline
TISI 	& $550 \times 960$	& 10	& 5819	& 1505 	 \\ \hline 
ERCe	& $480 \times 640$	& 5		& 9387 	& 4430 	 \\ \hline 
\end{tabular}
\end{table}

\vspace{.1cm}
\noindent \textbf{Visual and non-visual sources} - 
We extracted the following set of visual features for representing visual content in each clip: 
(a) colour features including RGB and HSV; 
(b) local texture features based on Local Binary Pattern (LBP)~\cite{OjalaTPAMI2002}; 
(c) optical flow; 
(d) holistic features of the scene based on GIST~\cite{OlivaIJCV2001};
and (e) person and vehicle\footnote{No vehicle detection on the ERCe dataset.} detection~\cite{FelzenszwalbGMR_PAMI10}.

We collected $10$ types of non-visual sources for the TISI dataset:
(a) weather data extracted from the WorldWeatherOnline with $9$ elements: temperature, weather type, wind speed, wind direction, precipitation, humidity, visibility, pressure, and cloud cover;
(b) traffic speed data from the Google Maps with $4$ levels of traffic speed: very slow, slow, moderate, and fast.
For the ERCe dataset, 
we collected data from multiple independent on-line sources 
about the time table of campus events including: 
No Scheduled Event (No Schd. Event), Cleaning, Career Fair, 
Gun Forum Control and Gun Violence (Gun Forum), Group Studying, 
Scholarship Competition (Schlr. Comp.), Accommodative Service (Accom. Service), 
Student Orientation (Stud. Orient.).

Note that other visual features and non-visual data types 
can be considered without altering the training and inference methods of our model 
in that the MSC-Forest model is capable of coping with different families of visual features 
as well as distinct types of non-visual sources.

\vspace{0.1cm}
\noindent \textbf{Baselines} - 
To evaluate the proposed method for multi-source video
clustering and tag inference, we compared the Visual + Non-Visual + MSC-Forest
(\textit{VNV-MSC-Forest}) model against the following baseline models:

\begin{enumerate}
\item 
\textit{VO-Forest}: 
a conventional forest~\cite{BreimanML01} trained with visual feature vectors alone, 
to demonstrate the benefits from using non-visual sources\footnote{ 
Evaluating a forest that takes only non-visual inputs is not possible, since non-visual
data is not available for previously-unseen video footages.}.

\item 
\textit{VNV-Kmeans}: 
$k$-means using concatenated vectors of visual and non-visual features, 
to highlight the heteroscedasticity and dimensionality discrepancy problem 
caused by heterogeneous visual and non-visual data.

\item 
\textit{VNV-Forest}: 
a conventional forest~\cite{BreimanML01} trained 
with concatenated visual and non-visual feature vectors, 
to compare the effectiveness of MSC-Forest 
that exploits non-visual data during forest formation.

\item 
\textit{VNV-AASC}:
a state-of-the-art multi-source spectral clustering method~\cite{Huang2012} learned by treating each type of visual or non-visual feature as an individual source, 
to demonstrate the superiority of MSC-Forest 
in handling diverse data representations and 
correlating multiple sources.

\item 
\textit{VNV-MSC-Forest-hard}:
a variant of our model using hard cluster assignment strategy for  
inferring semantic tags of previously-unseen samples (Section~\ref{sec:method@tagging}), 
to highlight the effectiveness of the proposed tree structure based tag inference algorithm.

\item
\textit{VT-MSC-Forest}:
a variant of our model using only temporal information and visual data. 
In order to show the exact effectiveness of exploiting non-visual data,
the weight ratio between visual data and time retains the same as in VNV-MSC-Forest
with the only difference of discarding non-visual data during model training.

\item 
\textit{VPNV$\rho$-MSC-Forest}:
a variant of our model but with $\rho$\% of training samples 
having arbitrary number of missing non-visual types, 
to evaluate the robustness of MSC-Forest 
in coping with partial/missing non-visual data.
\end{enumerate}

\vspace{0.1cm}
\noindent \textbf{Implementation details} - 
The clustering forest size $T_{\mathrm{clust}}$ was set to $1000$, 
including both the conventional forest and the proposed MSC-Forest.
We observed a slight increase in performance given a larger forest size, which agrees with~\cite{criminisi2012decision}.
The training set $\mathbf{X}_t$ of the $t$th MSC-tree was obtained 
by performing random selection 
with replacement from the augmented data space (Fig.~\ref{fig:view_RF}(b)).
We set $m_\mathrm{try} = \sqrt{d}$ 
with $d$ the data feature dimension (Eqn.~(\ref{eqn:split_parameter_optimisation})).
This is typically practiced~\cite{BreimanML01}.
We employed linear data separation~\cite{criminisi2012decision} 
as the test function for node splitting. 
We set the same number of clusters across all methods.
This cluster number was discovered automatically using the method presented in~\cite{ZelnikNIPS2004}.
For each dataset, $\sim75\%$ out of the total data was utilised for model training, and the remaining was reserved for testing.
Additional previously-unseen video data was collected from the Time Square Intersection scene on a separate day for video summarisation.


\section{Evaluations}
\label{sec:Exp}

\subsection{Multi-Source Clustering}
\label{sec:Exp@cluster_discovery}

To evaluate the effectiveness of different clustering models for
multi-source video clustering,
we compared the quality of their clusters formed on the training dataset.
%
For determining clustering quality, 
we quantitively measured the mean entropy~\cite{zhao2004empirical} of 
non-visual distributions $p(y_i|c)$ (Eqn.~(\ref{eqn:nonvisual_data_pdf})) 
associated with training data clusters
to evaluate how coherent video content are partitioned,
assuming all methods have access to non-visual data during the entropy computation.

\begin{figure*}[t]
\centering
\includegraphics[width=1\linewidth]{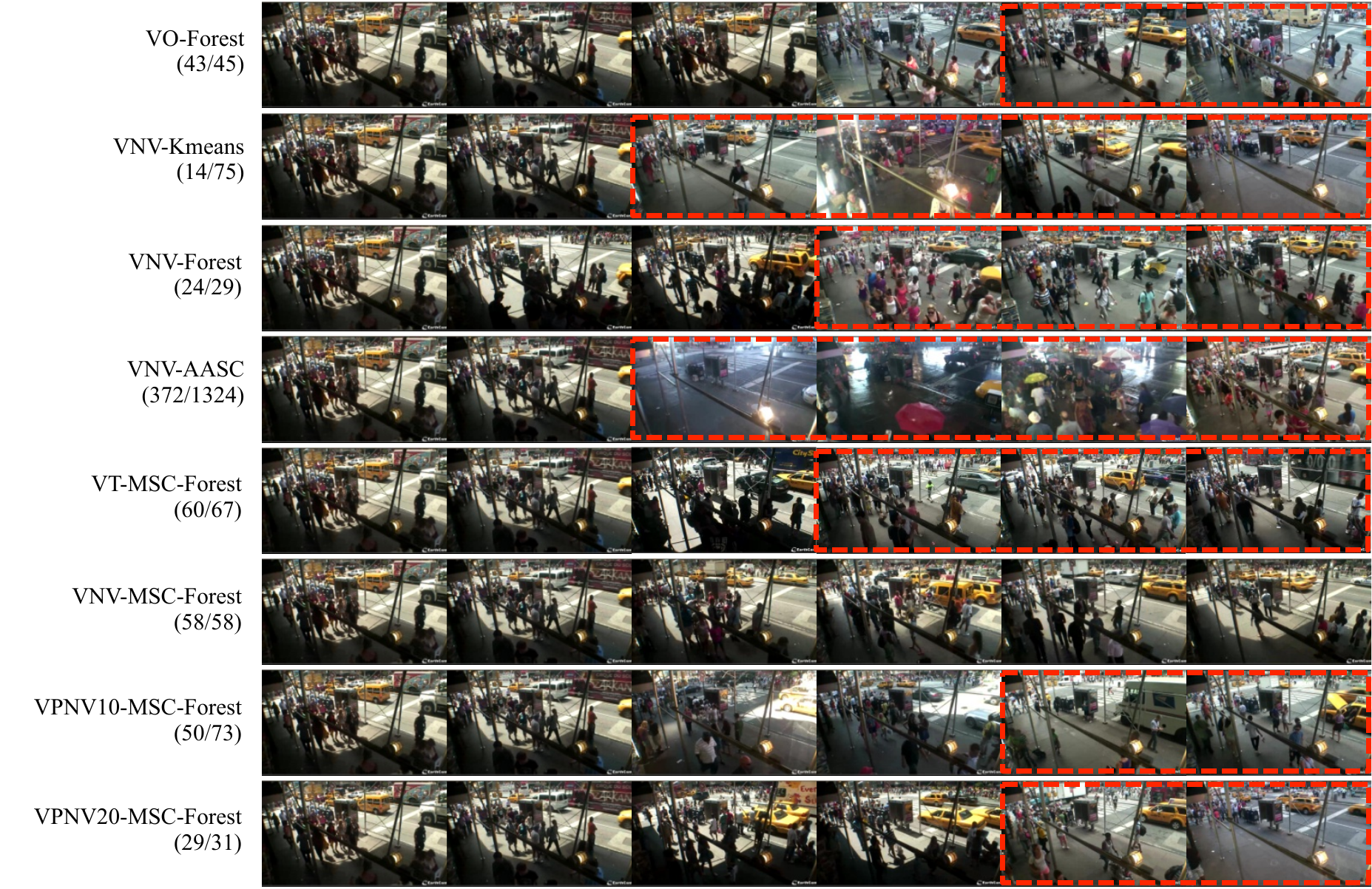}
\caption{\footnotesize 
Qualitative comparison on cluster quality on TISI.
A key frame of each video is shown. 
(X/Y) in brackets: 
X = the number of clips with sunny weather;   
Y = the total number of clips in a cluster. 
The frames inside the red boxes are inconsistent clips in a cluster.
}
\label{fig:cluster_eg_TISI}
\end{figure*}

\begin{table} 
\caption{
Compare cluster purity in mean entropy. Lower is better.}
\label{table:cluster_purity_mean_entropy}
\centering
\begin{tabular}{l|c|c|c} 
	Dataset					& \multicolumn{2}{|c|}{TISI} & ERCe \\ \hline
	$p(\mathbf{y}|c)$     	& traffic speed & weather & event \\ \hline \hline
\text{VO-Forest}			& 0.8675 & 1.0676 & 0.0616 \\ \hline
\text{VNV-Kmeans}			& 0.9197 & 1.4994 & 1.2519 \\ \hline
\text{VNV-Forest}			& 0.8611 & 1.0889 & 0.0811 \\ \hline
\text{VNV-AASC}				& 0.7217 & 0.7039 & 0.0691 \\ \hline
\text{VT-MSC-Forest}	    	& 0.7275 & 0.9577 & {0.0580}\\ \hline		
\text{VNV-MSC-Forest}	    & 0.7262 & \textbf{0.6071} & \textbf{0.0024}\\ \hline	
\text{VPNV10-MSC-Forest}	& \textbf{0.7190} & 0.6261 & \textbf{0.0024} \\ \hline	
\text{VPNV20-MSC-Forest}	& 0.7283 & 0.6497 & 0.0090 \\ \hline
\end{tabular}
\end{table}

It is evident from Table~\ref{table:cluster_purity_mean_entropy} that
our VNV-MSC-Forest achieves the best cluster purity on both datasets\footnote{
{VNV-MSC-Forest-hard} shares the same clusters as {VNV-MSC-Forest}.}.
Despite that there are gradual degradations in clustering quality when we increase
the non-visual data missing proportion, 
overall the {VNV-MSC-Forest} model copes well with partial/missing non-visual data.
With no aid of non-visual tag information, VT-MSC-Forest forms much worse clusters.
Whilst the superiority of VT-MSC-Forest over VO-Forest suggests the effectiveness of temporal information with MSC-Forest.
Inferior performance of {VO-Forest} to 
{VNV-MSC-Forest} suggests the importance of learning from auxiliary non-visual sources.
Nevertheless, not all methods perform equally well when learning from
the same visual and non-visual sources: the $k$-means and AASC perform
much poorer in comparison to MSC-Forest. 
The results suggest the proposed joint information gain criterion (Eqn.~(\ref{eqn:info_gain_our}))
is more effective in handling heterogeneous data 
than the conventional clustering models.

For qualitative comparison, we show examples in Fig.~\ref{fig:cluster_eg_TISI} using the
TISI dataset for detecting `sunny' weather.
It is evident that only \text{VNV-MSC-Forest} is able to provide
coherent video grouping, with only slight decrease in clustering
purity given partial/missing non-visual data. 
Other methods including \text{VNV-AASC} result in a large cluster
either leaving out some relevant clips or including many non-relevant ones, 
with most of them under the influence of strong artificial lighting sources. 
These non-relevant
clips are visually `close' to sunny weather, but semantically not. 
The VNV-MSC-Forest model avoids this mistake by correlating both visual and non-visual sources in an information theoretic sense.

%

\subsection{Video Tagging}
\label{sec:Exp@tagging}

Generating video summary with semantical interpretations 
requires accurate tag prediction.
In this experiment we compared the performance of different methods 
in inferring semantic tags given previously-unseen clips extracted from long videos. 
The proposed tagging algorithm (Section~\ref{sec:method@tagging}) is used 
for {VO-Forest}, {VT-MSC-Forest}, {VNV-MSC-Forest}, and {VPNV10/20-MSC-Forest}, whilst nearest neighbour (NN) strategy for the others. 
%
For quantitative evaluation, we manually annotated $3$
weather conditions (sunny, cloudy and rainy) and 
$4$ traffic speeds on TISI previously-unseen clips, 
whilst $8$ event categories on ERCe previously-unseen clips.

\begin{figure*}[t]
\centering
\includegraphics[width=0.6\linewidth]{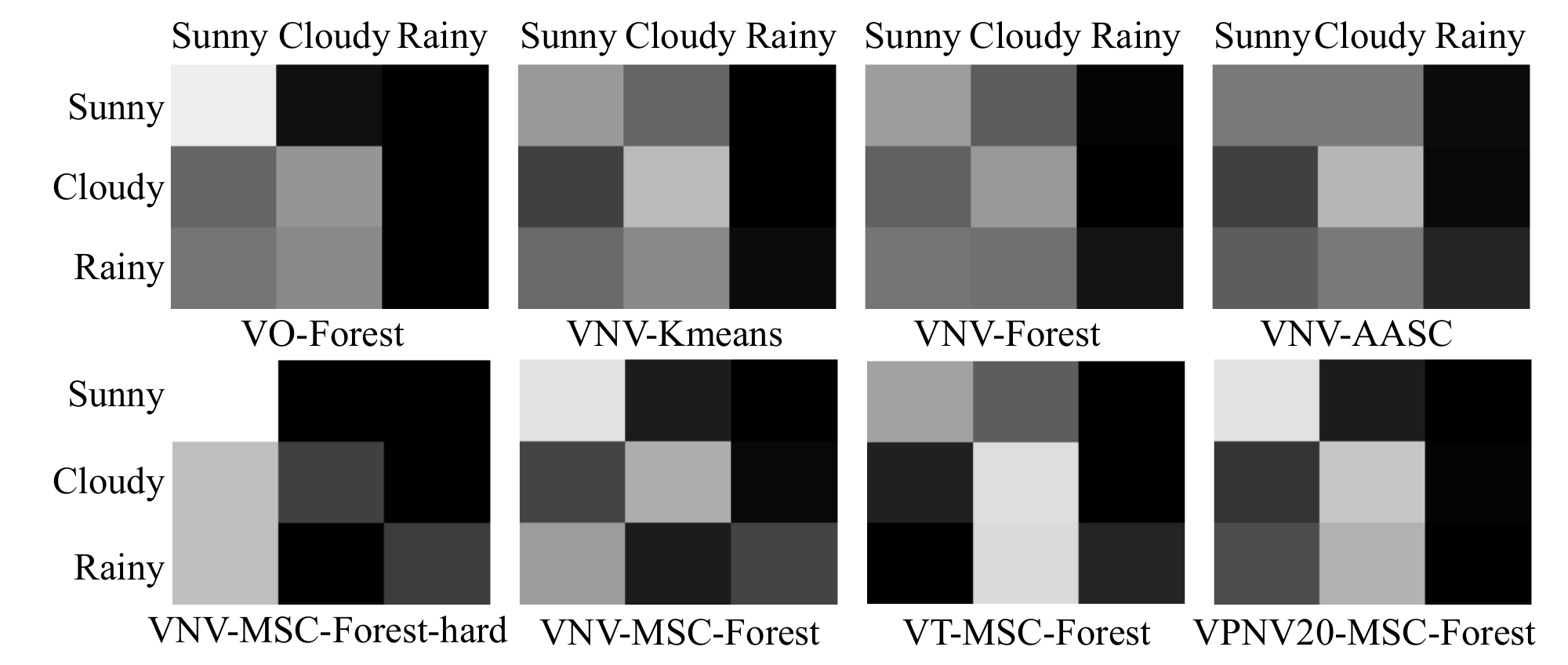}
\caption{\footnotesize Weather tagging confusion matrices (TISI dataset).}
\label{fig:confusion_mat_TISI}
\end{figure*}

\vspace{.1cm}
\noindent \textbf{Tagging video by weather and traffic conditions} -  
The experiment was
conducted on the TISI outdoor dataset. 
It is observed that the performance of different methods
(Table~\ref{table:tagging_acc_TISI}) is largely in line with 
their performance in data clustering 
(Section~\ref{sec:Exp@cluster_discovery}).
The poorest result of tagging traffic conditions is yielded by {VO-Forest}.
This suggests the significance of exploiting non-visual data during model training.
%
It is also seen from Fig.~\ref{fig:confusion_mat_TISI} that
{VNV-MSC-Forest} not only outperforms
other baselines in isolating the sunny weather, 
but also performs well in distinguishing visually ambiguous cloudy and rainy weathers. 
In contrast, both {VNV-Kmeans} and {VNV-AASC} mistake most
of the `rainy' scenes as either `sunny' or `cloudy', as they can be
visually similar.

\begin{table} \footnotesize 
\caption{\footnotesize Comparison of tagging accuracy on the TISI dataset.}
\label{table:tagging_acc_TISI}
\centering
\begin{tabular}{l|c|c} 
(\%)               & traffic speed & weather \\ \hline \hline
\text{VO-Forest}           & 27.62 & 50.65 \\ \hline
\text{VNV-Kmeans}          & 37.80 & 43.14 \\ \hline
\text{VNV-Forest}          & 34.95 & 43.81 \\ \hline
\text{VNV-AASC}            & 36.13 & 44.37 \\ \hline
\text{VNV-MSC-Forest-hard}  & 32.86 & 49.59 \\ \hline
\text{VT-MSC-Forest}        & 35.99 & {54.47} \\ \hline
\text{VNV-MSC-Forest}       & 35.77 & \textbf{61.05} \\ \hline
\text{VPNV10-MSC-Forest}    & 37.99 & 55.99 \\\hline
\text{VPNV20-MSC-Forest}    & \textbf{38.05} & 54.97 \\ \hline
\end{tabular}
\end{table}

\begin{table*}  \scriptsize 
\caption{\footnotesize Comparison of tagging accuracy on the ERCe dataset.}
\label{table:tagging_acc_ERCe}
\centering
\begin{tabular}{c|c|c|c|c|c|c|c|c|c} \hline
(\%) & 
\fontsize{2mm}{3mm}\selectfont{VO-Forest} & 
\fontsize{2mm}{2mm}\selectfont{VNV-Kmeans} & 
\fontsize{2mm}{2mm}\selectfont{VNV-Forest} & 
\fontsize{2mm}{2mm}\selectfont{VNV-AASC} & 
\fontsize{2mm}{2mm}\selectfont{VNV-MSC-Forest-hard} & 
\fontsize{2mm}{2mm}\selectfont{VT-MSC-Forest} & 
\fontsize{2mm}{2mm}\selectfont{VNV-MSC-Forest} & 
\fontsize{2mm}{2mm}\selectfont{VPNV10-MSC-Forest} & 
\fontsize{2mm}{2mm}\selectfont{VPNV20-MSC-Forest}\\ \hline \hline
No Schd. Event	& 79.48 & \textbf{87.91} & 32.47 & 48.51 & 81.25 & 57.43 & 55.98 & 47.96 & 55.57 \\ \hline
Cleaning		& 39.50 & 19.33 & 30.25 & 45.80 & 41.60 & \textbf{70.17} & 41.28 & {46.64} & 46.22 \\ \hline
Career Fair		& 94.41 & 59.38 & 65.46 & 79.77 & 70.07 & 91.45 & \textbf{100.0} & \textbf{100.0} & \textbf{100.0} \\ \hline
Gun Forum		& 74.82	& 44.30 & 45.77 & 84.93 & 60.48 & 79.96 & 83.82 & \textbf{85.29} & \textbf{85.29} \\ \hline
Group Studying	& 92.97 & 46.25 & 41.25 & 96.88	& 84.22 & \textbf{99.22} & {97.66} & {97.66} & 95.78 \\ \hline
Schlr Comp.		& 82.74 & 16.71 & 33.15 & 89.40 & 82.88 & 90.08 & 99.46 & \textbf{99.73} & 99.59 \\ \hline
Accom. Service	& 0.00  & 0.00 & 13.70 & 21.15 & 10.82 & 0.00 & \textbf{37.26} & \textbf{37.26} & 37.02 \\ \hline
Stud. Orient.	& 60.94 & 9.77 & 33.59 & 38.87 & 47.85 & 43.75 & 88.09 & \textbf{92.38} & 88.09 \\ \hline 
\hline
\textit{Average}& 65.61 & 35.45 & 36.96 &63.16 & 59.89 & 66.50 & 75.69 & 75.87 & \textbf{75.95} \\ \hline
\end{tabular}
\end{table*}

\begin{figure*}
\centering
\includegraphics[width=0.8\linewidth]{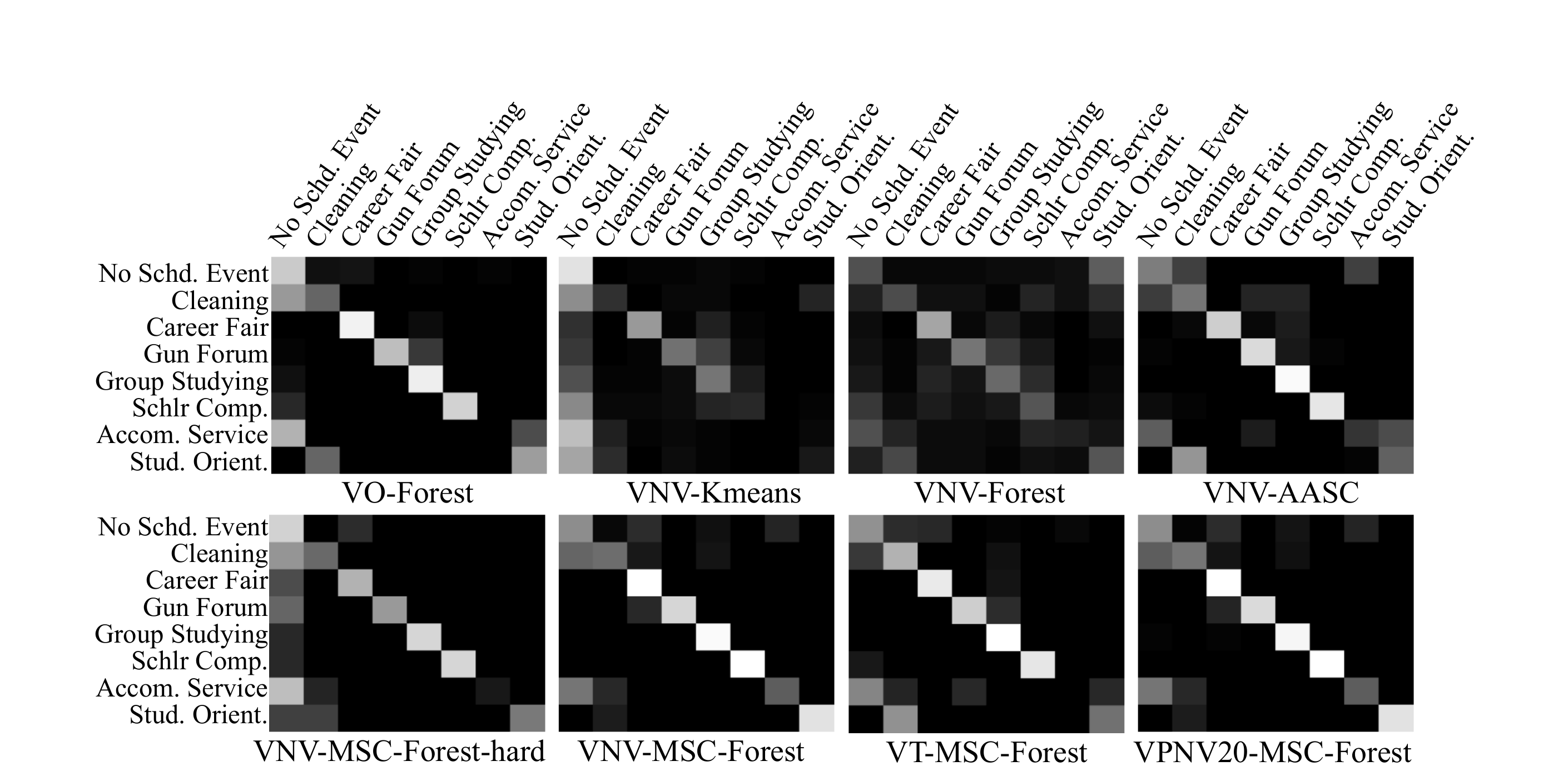}
\caption{\footnotesize Event tagging confusion matrices (ERCe dataset).}
\label{fig:confusion_mat_ERCe}
\end{figure*}

\vspace{.1cm}
\noindent \textbf{Tagging video by activity events} - 
Tagging semantic events was tested using the ERCe dataset. 
%
By {VO-Forest}, poor results (Table~\ref{table:tagging_acc_ERCe} and
Fig.~\ref{fig:confusion_mat_ERCe}) are obtained especially on
Accom. Service, which involves subtle activity patterns, 
\ie~students visiting particular rooms, 
suggesting using visual data alone is not sufficient to detect such events.
VT-MSC-Forest over-fits to 'Cleaning' event, therefore performs poorly on 'Stud. Orient' event.

Due to the typical high-dimension of visual sources 
compared to non-visual data,  
the latter is often overwhelmed by the former in representation.
{VNV-Kmeans} severely suffers from this problem 
as its most predictions are biased to No Schd. Event 
that is more common and frequent visually.
This suggests that this distance-based clustering is poor
in handling the heteroscedasticity and dimension
discrepancy problems in learning heterogeneous data.
{VNV-AASC} attempts to circumvent these problems by seeking for
an optimal combination of affinity matrices derived
independently from distinct data sources. However this is proved
challenging, particularly when each source is inherently noisy and
inaccurate.
%
In contrast, the proposed MSC-Forest correlates different
sources via a joint information gain criterion to effectively alleviate
these problems, 
leading to more robust and accurate tagging performance.
Again, {VPNV10/20-MSC-Forest} perform
comparably to {VNV-MSC-Forest}, further validating the robustness
of MSC-Forest in tackling partial/missing non-visual data
with the proposed adaptive weighting mechanism (Section~\ref{sec:method@adaptive_weighting}).

Interestingly, in some cases, {VPNV10/20-MSC-Forest} models
even outperform {VNV-MSC-Forest} slightly.
We observe that this can be caused by missing noisy non-visual data,
which may lead to better results.
Overall, the performance difference is marginal 
and the results demonstrate that MSC-Forest provides stable tagging results across both datasets.

\begin{figure}[h]
\centering
\subfigure [TISI.]
{
	\includegraphics[width=.45\linewidth]{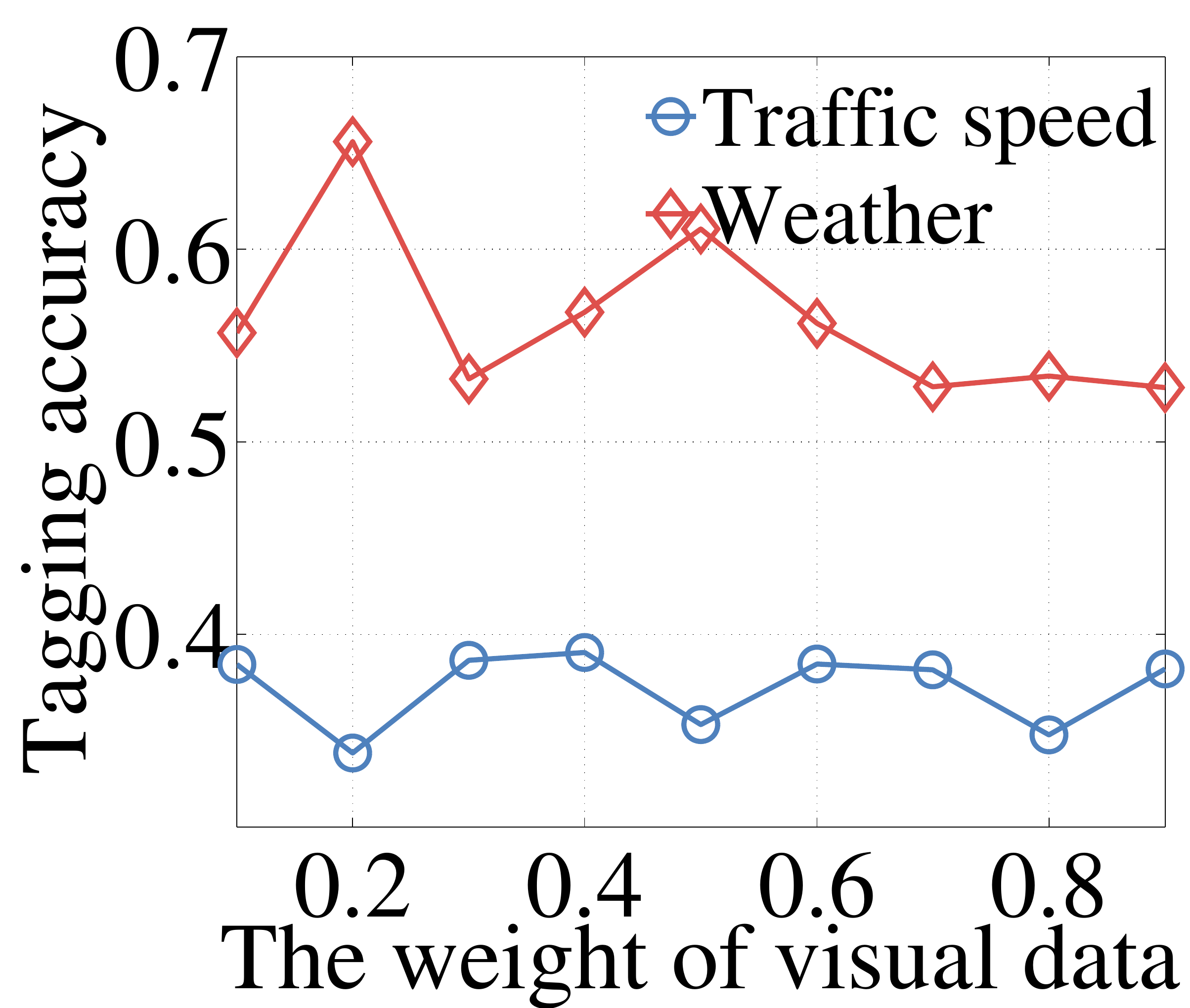}
}
\subfigure [ERCe.]
{
	\includegraphics[width=.45\linewidth]{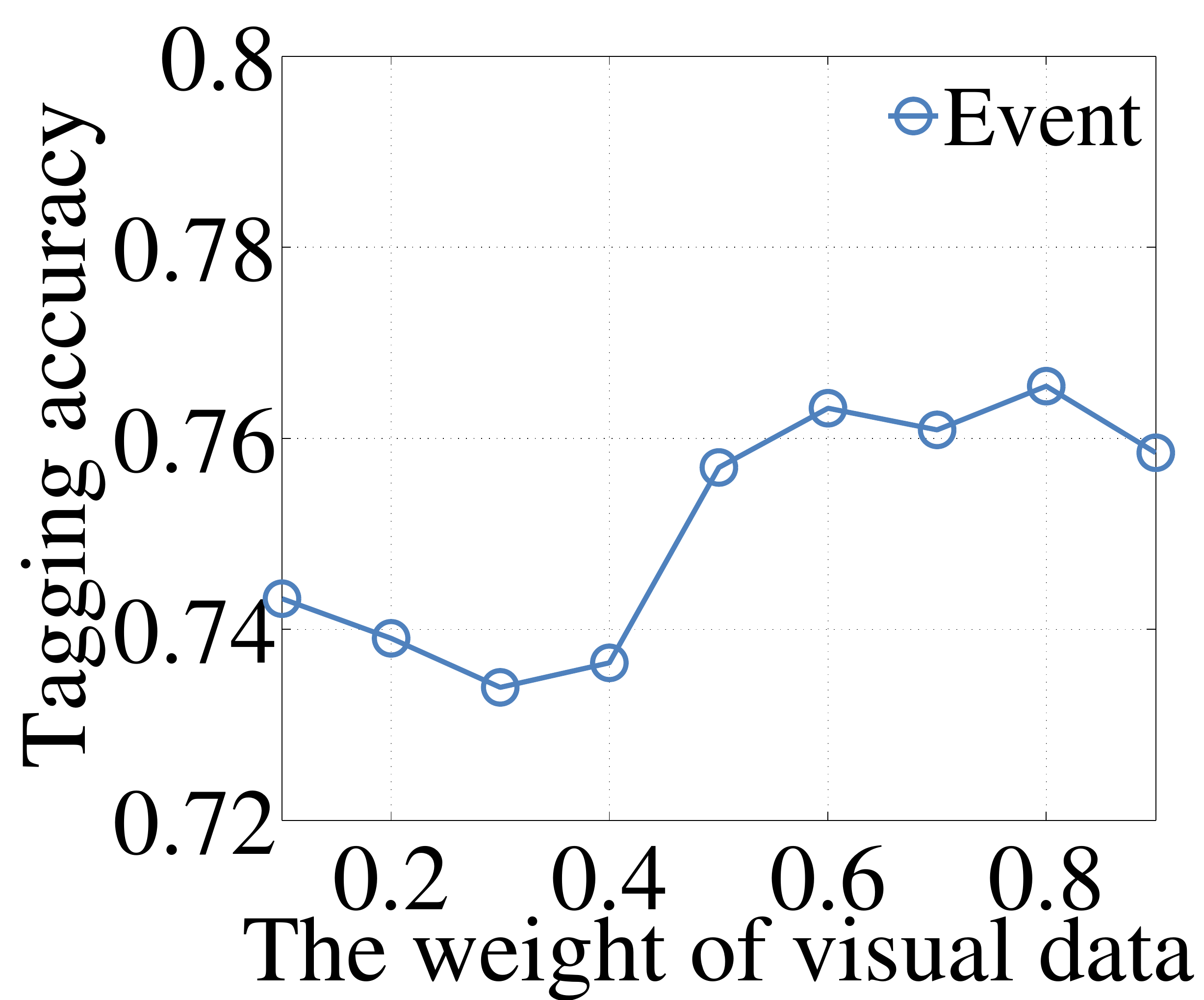}
}
\caption{ The average tagging accuracy against varying visual data weight $\alpha_v$ in Eqn.~(\ref{eqn:info_gain_our}).}
\label{fig:alpha}
\end{figure}

\vspace{0.1cm}
\noindent \textbf{$\alpha$ sensitivity} - 
We analyse the relative significance of visual data 
against non-visual and temporal data 
by varying its weight $\alpha_v$ (Eqn.~(\ref{eqn:info_gain_our})) 
in MSC-Forest during model training. 
The average tagging accuracy is utilised as performance measure criterion.
It is observed from Fig.~\ref{fig:alpha} that 
setting $\alpha_v=0.5$ achieves satisfactory results for both datasets.
This observation suggests that visual and non-visual data are almost equally informative. This setting of $\alpha$ is adopted throughout our experiments.

\subsection{Semantic Video Summarisation}
\label{sec:Exp@synopsis}

%
In this experiment, we follow the method
described in Section~\ref{sec:method@summarisation}, and show that the
learned model MSC-Forest can be easily extended to produce compact yet
meaningful video summary of previously-unseen video footage,
\eg~from the Time Square Intersection scene,
with automatically generated semantic tags.
Despite captured from the same scene as the TISI dataset, 
this previously-unseen video is challenging in that it contains a number of
events not seen before (\eg~scaffolding event), with very different
weather and traffic conditions. It is interesting to examine how well
the multi-source model could generalise for drawing meaningful
summarisation given such unexpected disparities.

\subsubsection{A Quantitative Evaluation on Summary Quality}
\label{sec:Exp@synopsis_quantitative_evaluation}

Measuring the quality of video sumary quantitatively is non-trivial since there is no formal definition in the literature. In this study, we employ a \textit{coverage} metric -- an ideal summary should cover as many events of interest as possible\footnote{The event of interest is analogous to important objects/regions in~\cite{LeeCVPR2012}.}.
More precisely, given a video summary $\mathcal{V}$, its coverage is defined as 
$\widetilde{c} = \frac{N_\mathrm{covered}} {N_\mathrm{all}} \left( \frac{\max_i{|\mathcal{V}_i|}}{|\mathcal{V}|} \right)$, where $N_\mathrm{covered}$ and $N_\mathrm{all}$ represent the number of covered and all events of interest, respectively.
The $|\mathcal{V}|$ is the length of the current summary, whilst $\max_i{|\mathcal{V}_i|}$ represents the maximum length of all comparative synopses. The term $\left( \frac{\max_i{|\mathcal{V}_i|}}{|\mathcal{V}|} \right)$ thus penalises a summary with longer length.
Higher coverage is better, implying lower redundancy.

\begin{figure*}[t]
\centering
	\includegraphics[width=.8\linewidth]{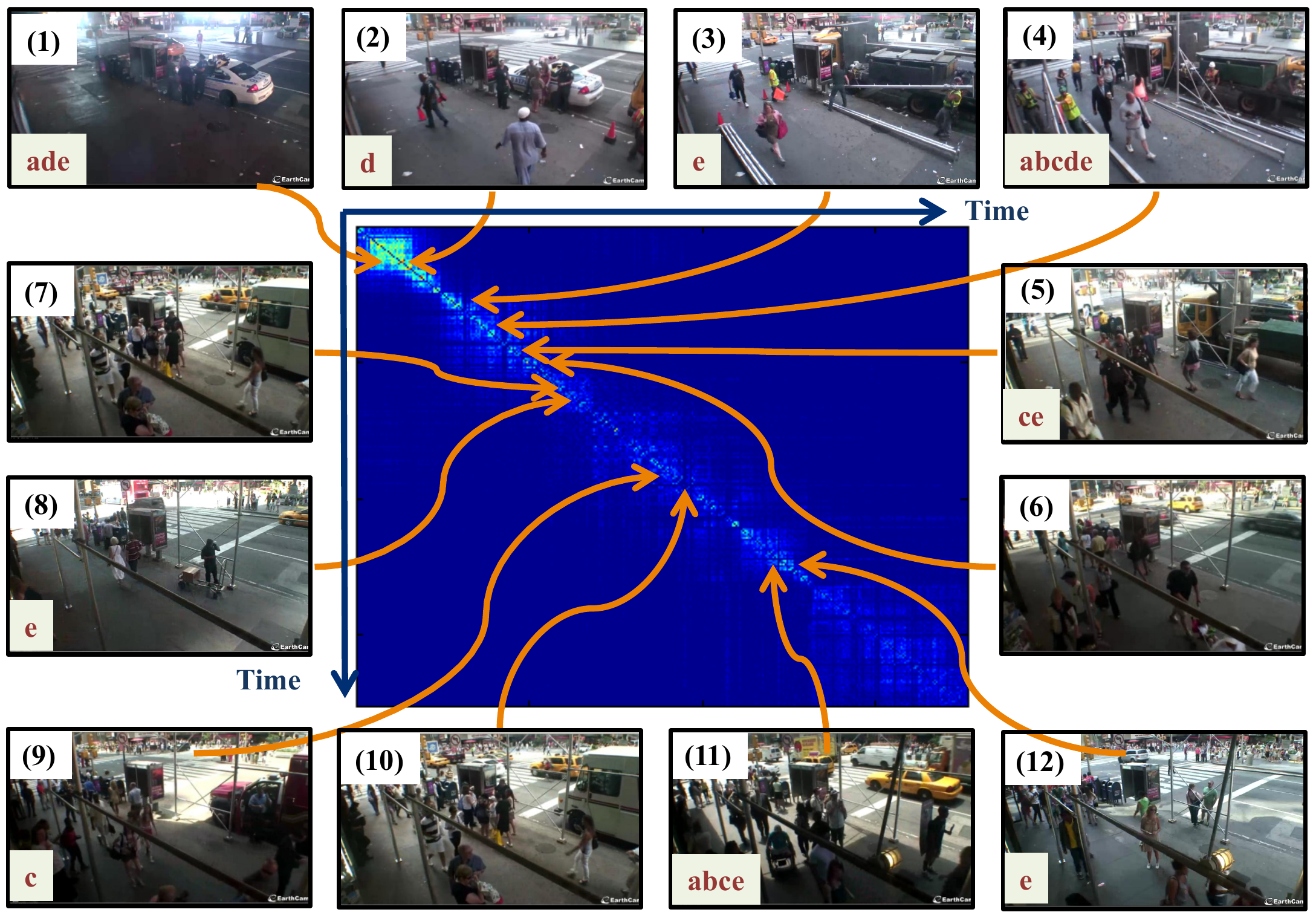}
\caption{\footnotesize The multi-source affinity matrix constructed by our model, along with key frames corresponding to ground truth events of interest: 
(1) policemen on-duty,
(2) blocking pathway,
(3) workers unloading scaffolding tubes,
(4)-(6) different stages of scaffolding,
(7)(9)(10) van parking aside,
(8) parcel delivery,
(11)(12) loitering events.
The event covered by some particular method is indicated on the left-bottom corner of key frame with their ID defined as:
(a) Uniform-Sampling;
(b) Sufficient-Change(L$1$);
(c) Sufficient-Change(L$2$);
(d) VO-Forest;
(e) {VNV-MSC-Forest}.
}
\label{fig:multi-source-aware_graph}
\end{figure*}

In order to generate unbiased ground truth of event of interest, we
asked $10$ annotators to watch the previously-unseen video carefully and label each video clip with arbitrary event tags.
Although these event tags were produced independently in a somewhat subjective manner, the repetition of similar tagging among different annotators is high, \eg~most annotators labelled `unloading scaffolding tubes', `policemen on-duty', as events of their interest. Thus, we formed the ground truth with events that were agreed by over $50$\% of the annotators.
The final ground truth consists of $12$ events (Fig.~\ref{fig:multi-source-aware_graph}).

Given the ground truth, we compared the quality of summary generated using the proposed multi-source MSC-Forest with the baselines: 
(1) Uniform-Sampling: 
a straightforward way of summarising video by
uniformly sampling video clips over time,
assuming key events are distributed evenly~\cite{truong2007video,LeeCVPR2012}.
(2) Sufficient-Change: a classical summarisation strategy generic to video
category~\cite{ZhangPR1997,kim2002object,truong2007video}.
The idea is to select the clip significantly different from the previous key clip
e.g. using threshold based strategy and 
thus the extracted key clips may be of great diversity and complete.
The threshold can be estimated based on the number of key clips.
For the distance metric, we adopt L$1$-norm and L$2$-norm 
to measure pairwise similarity between clips in our experiment.
(3) VO-Forest: the conventional Forest~\cite{BreimanML01} that exploits visual features alone.
For VO-Forest and MSC-Forest, we applied the summarisation pipeline described in Section~\ref{sec:method@summarisation} for summary composition.
We generated the video summary by the remaining methods via setting a duration
similar to the summary by MSC-Forest.
Note that non-visual information are not available during the
summarisation stage. Hence, for clustering based models,
the quality of a summary essentially ties
to the purity and coherency of video clusters discovered using different methods.
%

The results are shown in Fig.~\ref{fig:multi-source-aware_graph}
and Table~\ref{table:synopsis_quantitative_comparison}.
It is evident that the MSC-Forest model achieves higher event coverage
than the baselines. This is in large due to the
MSC-Forest's ability for latent data structure discovery
(Section~\ref{sec:Exp@cluster_discovery}). 
To reveal concrete reasons on the summarising performance difference,
for the same previously-unseen samples $\mathbf{x}^*$
with event of interest, \eg~parcel delivery, we compared the
assigned clusters: $c_{\mathrm{vnv}}^*$ by our model and $c_{\mathrm{vo}}^*$
by VO-Forest. 
It is found that samples in $c_{\mathrm{vnv}}^*$ are visually consistent
each other and the majority share some similarity with 
$\mathbf{x}^*$, \eg~someone standing at the edge of
pathway; whilst cluster $c_{\mathrm{vo}}^*$ is much larger with
no obvious visual commonality over its cluster members.
Uniform-Sampling performs poorly 
since the assumption of uniform event distribution is often invalid. 
Significant-Change is inferior to our model since the visual data distance/similarity measure can be inaccurate and less meaningful due to the challenging semantic gap problem.

\begin{table} [t] \footnotesize
\caption{ Quantitative comparison of summary. Length = clip number.}
\label{table:synopsis_quantitative_comparison}
\centering 
\begin{tabular}{c|c|c|c}
Method                  & Length & Event number & Coverage \\
\hline
\hline
Uniform-Sampling        	& 28 & 3 & 25.9\% \\ \hline
Sufficient-Change(L$1$) 	& 29 & 2 & 16.7\% \\ \hline
Sufficient-Change(L$2$) 	& 29 & 4 & 33.3\% \\ \hline
VO-Forest               	& 21 & 3 & 34.5\% \\ \hline
VNV-MSC-Forest(\textbf{Ours})& 28 & \textbf{7} & \textbf{60.4}\% \\ \hline
\end{tabular}
\end{table}

\subsubsection{A User Study on Summary Quality}
\label{sec:Exp@user_study}

\begin{figure*} [t]
\centering
	\includegraphics[width=1\linewidth]{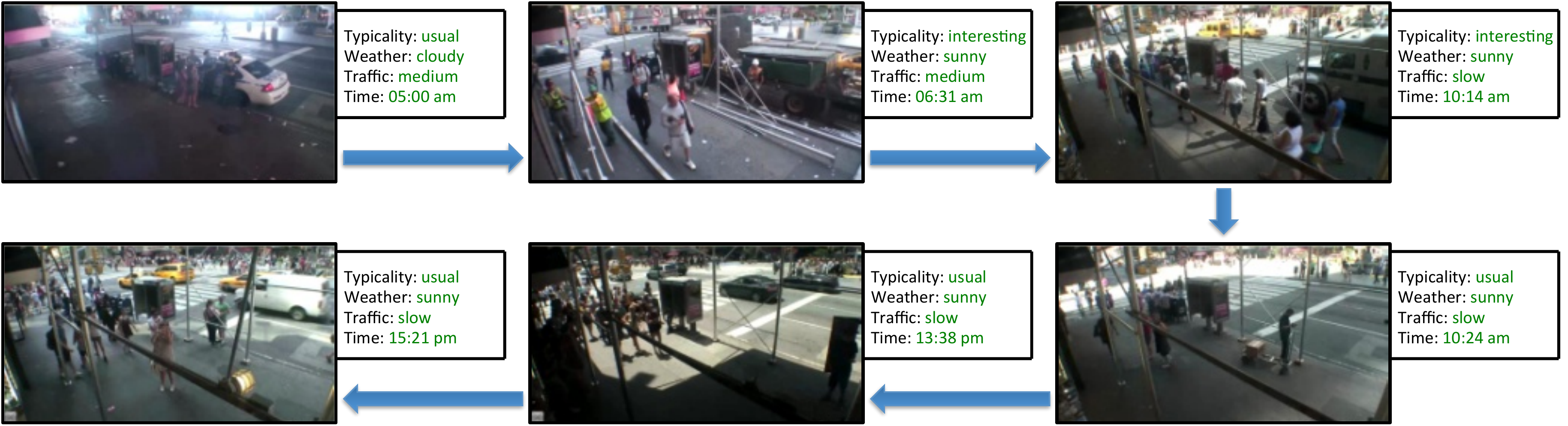}
\caption{\footnotesize A storyboard version of our video summary enriched with non-visual tags. 
}
\label{fig:video-synopsis_TISI}
\end{figure*}

We conducted a user study to examine if the non-visual tags inferred using the MSC-Forest model could complement the unilateral perspective offered by pure visual summary alone.
We showed two video summaries to $10$ volunteers: (i) a pure visual summary, and (ii) the same summary but enriched with semantic tags inferred using the proposed multi-source model\footnote{The inferred non-visual tags include weather, traffic conditions, and typicality. The typicality tag, \ie~\textit{usual} and \textit{interesting}, of each clip, is computed based on the size of their assigned clusters (Fig.~\ref{fig:keyclips_extraction}(c)).
Clips assigned to the top $20\%$ smallest clusters are treated as `interesting'.}. The tagged summary is shown in Fig.~\ref{fig:video-synopsis_TISI}. Each volunteer was asked to compare and rate the two summaries based on their preference.  
It is worth pointing out that passing the user test is challenging 
because providing additional non-visual tags to summary is not necessarily better than none. 
Tags that correlate poorly with visual context could even jeopardise user experience.

\begin{figure} [t]
\centering
	\includegraphics[width=1\linewidth]{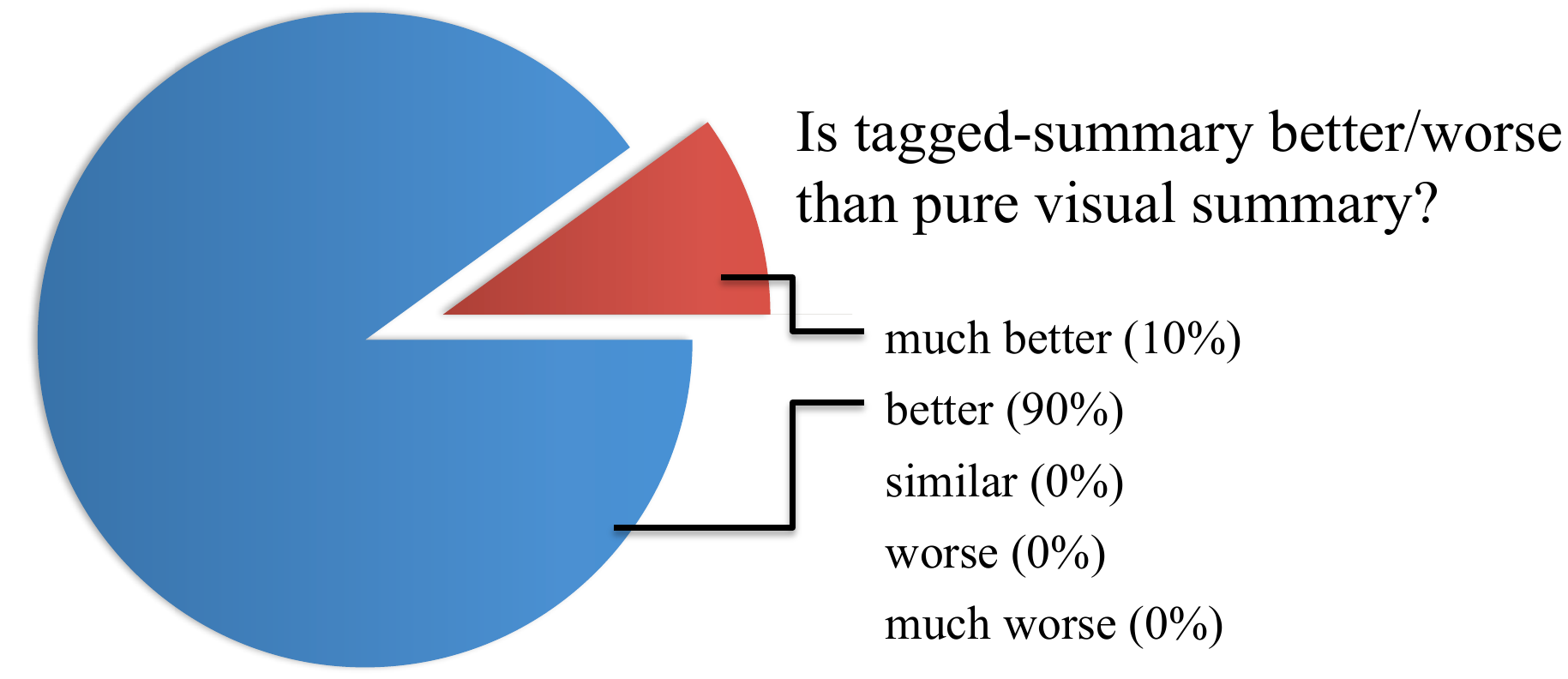}
\caption{\footnotesize User study:
 tagged \textit{versus}
 pure-visual summary.}
\label{fig:user_study}
\end{figure}

It is evident from Fig.~\ref{fig:user_study} that visual summary augmented with non-visual tags was well accepted by all participants over the conventional visual-only summary. 
A follow-up survey with the volunteers reveals several interesting reasons of their selection. 
Many volunteers found that the inferred non-visual tags were valuable
in providing auxiliary  context to achieve better global situational
awareness. In particular, the tags helped them to `connect the dots'
and making sense of the previously-unseen (and likely unfamiliar) video footages.
Some other volunteers credited the additional non-visual tags in
focusing their attention on particular events, and helping them in
spotting `outliers' of interest.

This user study provides an independent means to analyse and validate the usefulness
of visual summarisation with auto-tag inference of previously-unseen video
footages without a priori semantics or meta-data, mostly typical of
surveillance videos. It also shows the effectiveness of the proposed
model for mapping multi-source non-visual information to unstructured
and previously-unseen video data in automatic tagging and summarisation of the videos.

\subsection{Multi-Source Model Visualisation}
\label{sec:model_visualisation}
The superior performance of \text{VNV-MSC-Forest} 
can be better explained by examining more closely 
the capacity of MSC-Forest in
uncovering and exploiting the intrinsic correlation among different
visual sources and more critically among visual and non-visual sources.
This indirect correlation among heterogeneous sources 
results in well-structured decision trees, 
subsequently leading to more consistent data clusters 
and more accurate semantics inference.
The details of computing the multi-source correlation are presented in Appendix~\ref{sec:Exp_quantify_correlation}.
Here we show an example multi-source correction revealed by our MSC-Forest for model visualisation purpose.

%
%

\begin{figure}
\centering
\subfigure [ Visual-visual.]
{
	\includegraphics[width=.65\linewidth]{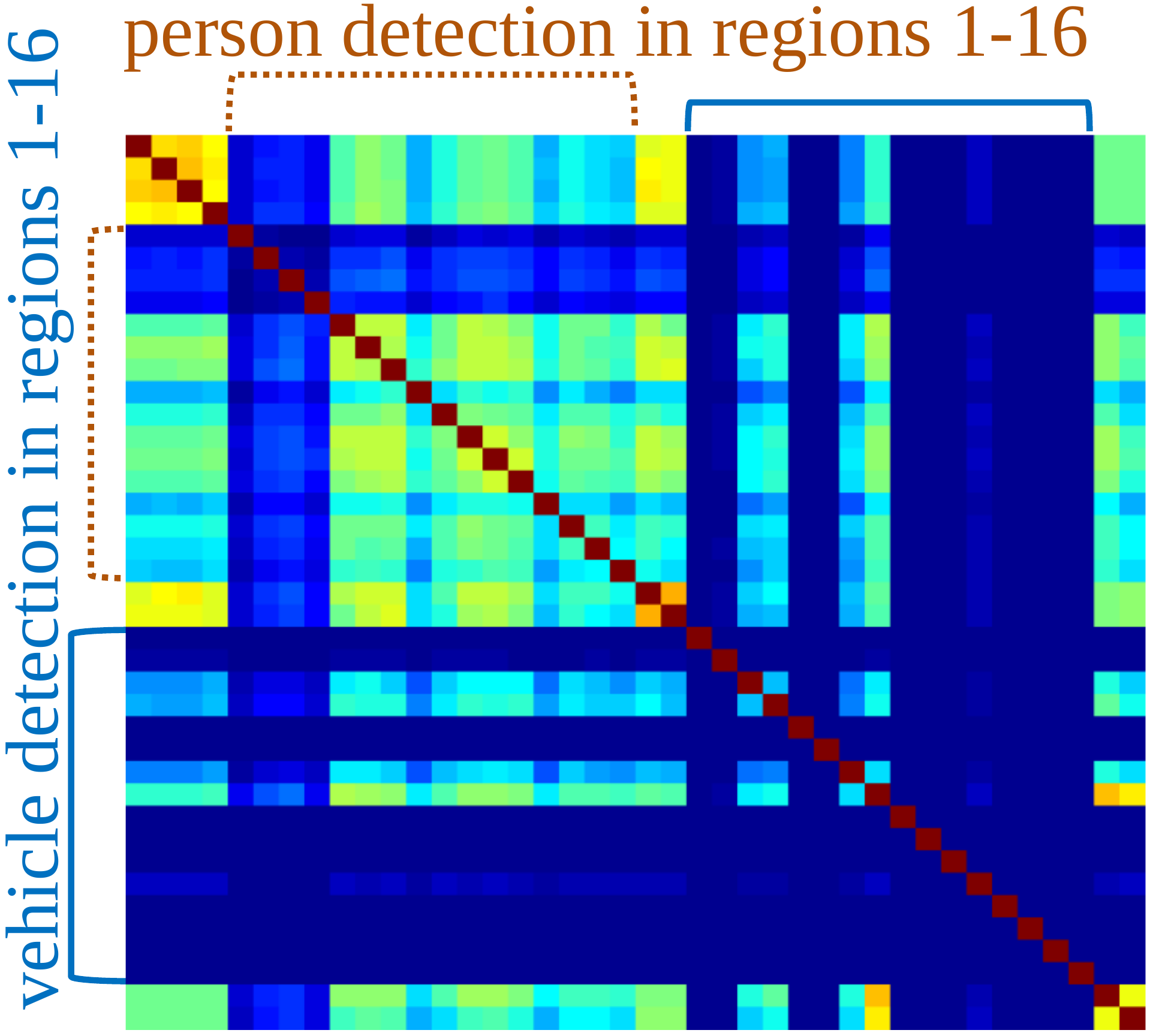}
}
\subfigure [ Vehicle detection and traffic speed.]
{
	\includegraphics[width=.7\linewidth]{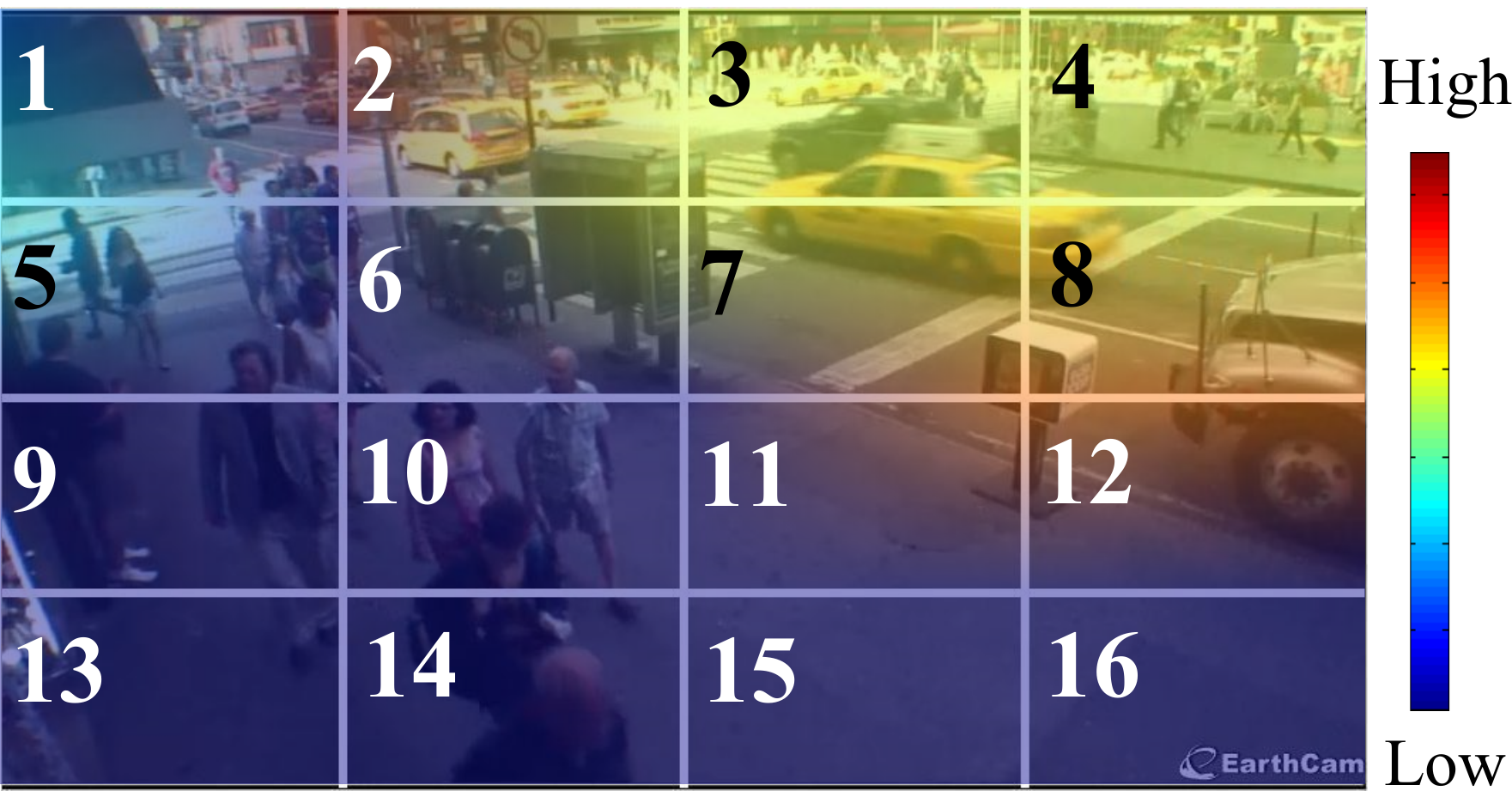}
}
\caption{The discovered multi-source correlation by our MSC-Forest on TISI.}
\label{fig:src_correlation}
\end{figure}

Intuitively, vehicle and person counts 
should correlate in a busy scene like TISI.
Our MSC-Forest discovered this correlation (see Fig.~\ref{fig:src_correlation}(a)), 
so the less reliable vehicle
detection from distance against a cluttered background, could enjoy a
latent support from more reliable person detection in regions $5$-$16$
close to the camera view.

Moreover, visual sources also benefit from correlated support
from non-visual data through 
our cross-sources information gain optimisation 
(Eqn.~(\ref{eqn:info_gain_our})).
An example is the intuitive correlation between traffic speed and visual appearance, e.g.
slow traffic speed often corresponds to crowded scenarios with a large quantity of
pedestrians and vehicles whilst fast traffic speed to sparse people and cars.
%
%
Such cross-source correlation can be captured by our MSC-Forest, 
as observed in Fig. 10(b) that 
the vehicle detection responses over road area present a stronger
interaction with traffic speed data than those on walk path 
where vehicles should not appear.
In other words, vehicle detection features of road area are preferred over those on walk path
in node splitting due to larger induced joint information gain (Eqn. (\ref{eqn:info_gain_our})),
which is clearly desired.
This discovered correlation is further exploited by MSC-Forest
during the node splitting optimisation process and 
thus facilitates the separation
of different crowdedness levels of visual data.
This leads to better clusters and eventually benefits video summarisation.

\subsection{Computational Costs and Model Complexity}
\label{sec:exp@complexity}

We examined the computational costs for
training the proposed MSC-Forest, in comparison to the
conventional forests. 
Time is measured on a Windows PC machine with a dual-core CPU @ 2.66 GHz, 4.0GB RAM, 
with C++ implementation. Only one core is utilised for training each forest.
%
We recorded the model training time under the same experimental setting as stated in Section~\ref{sec:Exp_Settings}.
It is observed from Table~\ref{table:forest_training_complexity} that the 
training cost of a MSC-Forest model is significantly lower than that of learning conventional forests.
In particular,
\text{VNV-MSC-Forest} records a reduced training time by
$14.4$\% and $17.1$\% on TISI, and
$64.1$\% and $64.4$\% on ERCe, when compared with \text{VO-Forest} and \text{VNV-Forest}, respectively.
We observed similar trend on the model inference time.

\begin{figure} [h]
\centering
\subfigure 
{
   	\includegraphics[width=.7\linewidth]{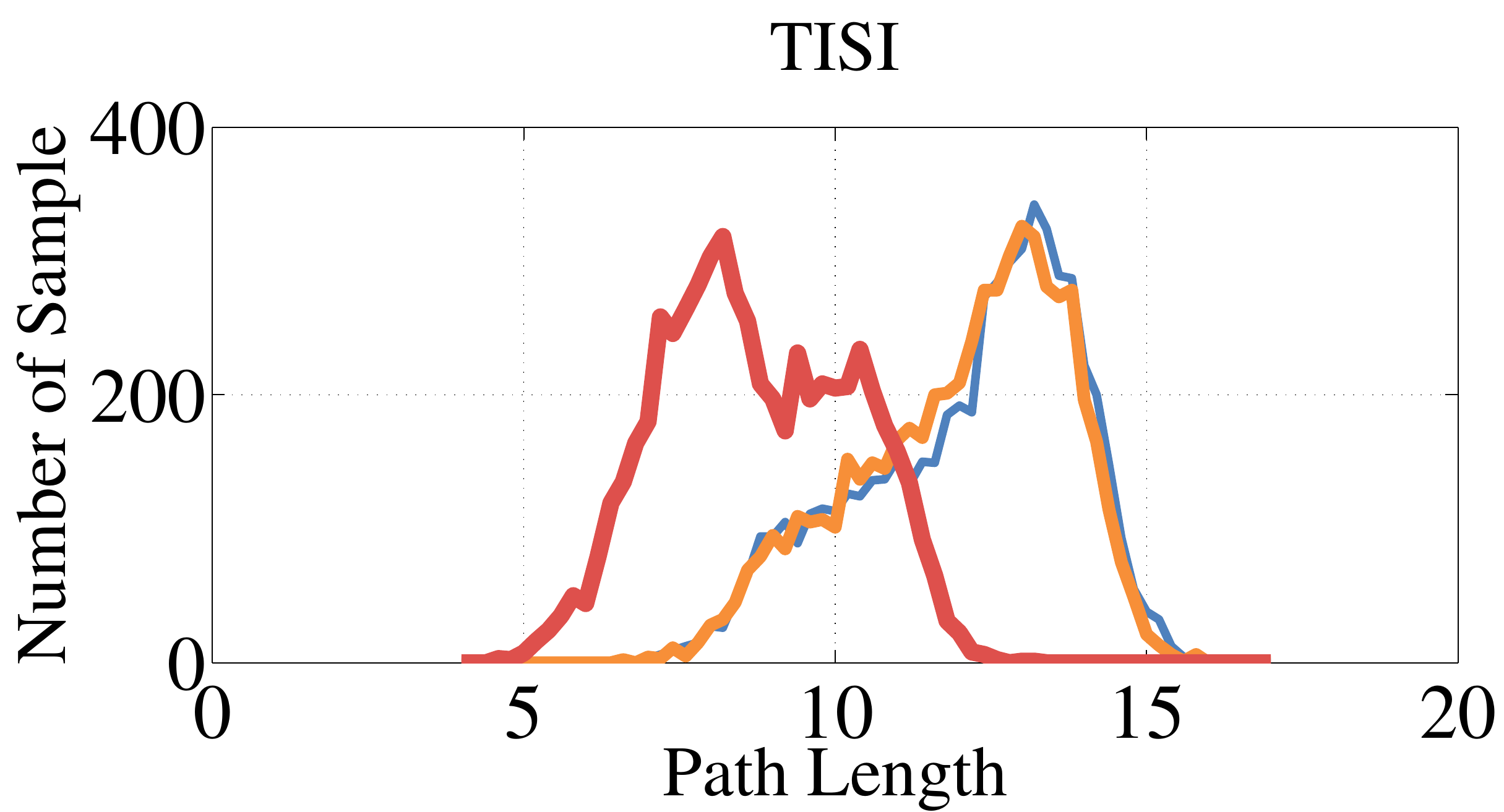}
}
\subfigure 
{
   	\includegraphics[width=.7\linewidth]{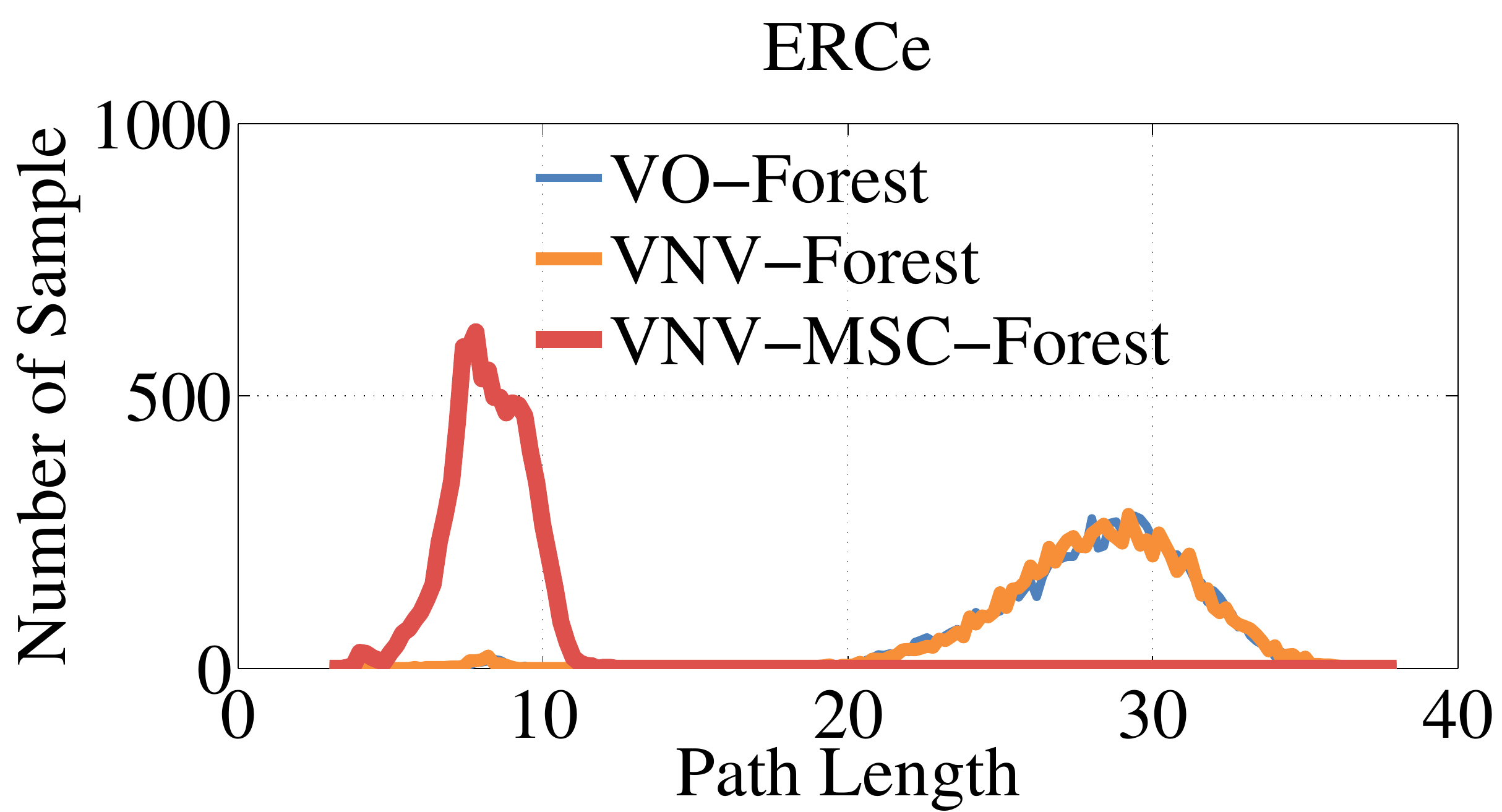}
}
\caption{ Comparing tree path length statistics.
   	The same legend is used for both charts.
   	}
\label{fig:tree_structure}
\end{figure}

The lower computational cost of MSC-Forest is owing to its shallow and balanced trees, thanks to the additional non-visual and temporal information during tree optimisation.
To make this concrete, we showed in Table~\ref{table:forest_training_complexity} the averaged tree fan-in
$\Phi^* = \frac{1}{T_\mathrm{clust}} \sum_t^{T_\mathrm{clust}} \Phi(t)$ 
of different forest models. A forest with shallow and balanced trees tend to have a small $\Phi^*$ (see Section~\ref{sec:method@complexity} for a discussion on tree fan-in).
In addition, we also profiled the length of path (from root to leaf node) traversed by training samples. A shallow and balanced tree tends to have shorter path length. The distributions depicted in Fig.~\ref{fig:tree_structure} suggest that MSC-Forest has a shallower and more balanced tree topology than that of conventional forests.
It is worth pointing out that despite the shallower structure, MSC-Forest outperforms other models in our clustering and tagging experiments.




\begin{table} [h] 
\caption{
Random forest model training complexity.
Lower is better. TT = Training Time (unit is second).}
\label{table:forest_training_complexity}
\centering
\begin{tabular}{l|c|c|c|c} 
Dataset  & \multicolumn{2}{c|}{TISI} & \multicolumn{2}{|c}{ERCe} \\ \hline 
    -                 & TT & $\Phi^*$ & TT & $\Phi^*$ \\ \hline \hline
{VO-Forest}    & 10306 & 109392 & 21831 & 359247 \\ \hline
{VNV-Forest}   & 10646 & 108865 & 22015 & 359364 \\ \hline
{VNV-MSC-Forest}& \textbf{8823} & \textbf{91316} & \textbf{7845} & \textbf{137620} \\ \hline
\end{tabular}
\end{table}

\section{Conclusion and Future Work}
\label{sec:conclusion}

We have presented a novel unsupervised multi-source learning model for
video summarisation.
Specifically, we introduced a joint information gain function for discovering and exploiting latent correlations among
independent heterogeneous data sources. 
The function naturally copes with diverse types of data with different representations, distributions, and dimensions.
Importantly, our model is capable of tolerating partial and missing
non-visual data, lending it well for automatic semantic tag
inference on previously-unseen video footages and for video summarisation.
Furthermore, the proposed joint optimisation encourages more compact decision trees, leading to more efficient model training and semantic tag inference.
Comparative experiments have demonstrated the advantages of the proposed
multi-source video clustering model over
existing visual-only models, for both discovering latent video
clusters and inferring non-visual semantic tags on previously-unseen
video footages. A comprehensive user
study was carried out to validate independently the
effectiveness of deploying the proposed model for generating
contextually-rich and semantically-meaningful video summary.

The proposed model is not limited to surveillance-type videos but can
be generalised to other types of unstructured and un-tagged consumer
videos or egocentric videos, if 3D camera motion-invariant features or
egocentric features~\cite{LeeCVPR2012} are adopted. 
%
%
For future work, we will consider generalising/transferring a learned model to new
scenes that are significantly different from the training environments. This can be partly addressed by utilising intermediate data representations such as attributes.

\appendices
%

\section{Quantifying Correlation between Sources}
\label{sec:Exp_quantify_correlation}
Quantifying latent correlation between different sources gives insights into their interactions in forming coherent video groupings.
This can be done once a MSC-Forest is trained.
To quantify between-source correlation, we first 
estimate correlation among their constituent features. 

\vspace{0.1cm}
\noindent \textbf{Visual-visual feature correlation} -
%
Visual-visual feature correlation 
is typically quantified based on their similarity in inducing split node partitions $L$ and $R$~\cite{BreimanML01}.
In particular, given a split node $s$ and its final optimal split,
say $L_{\nu}$ and $R_{\nu}$ by feature $\nu$. 
From Eqn.~(\ref{eqn:split_parameter_optimisation}),
we recall that this feature $\nu$ is selected out from the $m_\mathrm{try}$ randomly sampled features
$F^s=\{ f_1, \dots, f_{m_\mathrm{try}} \}$.
%
Let $\tau \in F^s \setminus \nu$ and 
its optimal left-right partitions be $L_{\tau}$ and $R_{\tau}$ respectively.
The node-level correlation between features $\nu$ and $\tau$ is then defined as
\begin{equation}
\lambda_f(\nu, \tau) = 
\frac{ p_{\nu} - (1 - 
\frac{|L_{\nu} \cap L_{\tau}|}{|L_{\nu} \cup R_{\nu}|}  - 
\frac{|R_{\nu} \cap R_{\tau}|}{|L_{\nu} \cup R_{\nu}|}) } 
{p_{\nu}},
\label{eqn:var_assoc_tree}
\end{equation}
where $p_{\nu} = \min(\frac{|L_{\nu}|}{|L_{\nu}|+|R_{\nu}|}, \frac{|R_{\nu}|}{|L_{\nu}|+|R_{\nu}|})$, thus $p_{\nu} \in (0, \frac{1}{2}]$.
With Eqn.~(\ref{eqn:var_assoc_tree}) we assign a strong correlation ($\lambda_f(\nu, \tau)=1$) 
to a feature pair ($\nu$, $\tau$) if they produce the same data partition, 
whilst a weak correlation ($\lambda_f(\nu, \tau) \leq -1$) 
when their partitions have no overlaps.
For simplicity we let
$\lambda_f(\nu, \tau) = \max(\lambda_f(\nu, \tau), 0)$
such that $\lambda_f(\nu, \tau)$ lies in the range of $[0, 1]$.
The final visual-visual feature correlation $\lambda(\nu, \tau)$ is obtained via
\begin{equation}
\lambda(\nu, \tau) = \frac{1}{T_\mathrm{clust}} \sum_{t=1}^{T_\mathrm{clust}} 
\left[ \frac{1}{N_{(\nu, \tau)}^t} 
\sum_k^{N_{(\nu, \tau)}^t} 
\lambda_f(\nu,\tau) \right],
\label{eqn:feature_corr_forest}
\end{equation}
where $N_{(\nu, \tau)}^t$ refers to the number of sampling co-occurrences of a feature pair 
($\nu$, $\tau$) during the splitting process of a MSC-tree $t$.

\vspace{0.1cm}
\noindent \textbf{Visual-nonvisual feature correlation} - 
Recall that visual and non-visual data play different roles in our MSC-Forest, 
e.g. the former as splitting features whereas the later as auxiliary information.
This difference makes the above equations not applicable to 
the computation of visual-nonvisual feature correlation
since no data split is associated with non-visual features.
%
%
Instead, we adopt information gain as the visual-nonvisual feature correlation metric.
This metric is appropriate in that it also reflects the intrinsic mutual interaction between visual and non-visual features during joint information gain optimisation (Eqn.~(\ref{eqn:info_gain_our})).
Formally, we quantify the node-level correlation between 
the optimal splitting visual feature $\nu$ and a non-visual feature $\omega$
as $\lambda_f(\nu,\omega) = \frac{\Delta \mathcal{I}_{\omega}}{\mathcal{I}_{\omega 0}}$ 
(the non-visual term of Eqn.~(\ref{eqn:info_gain_our})). 
The final visual-nonvisual feature correlation 
$\lambda(\nu, \omega)$ is computed similarly by Eqn.~(\ref{eqn:feature_corr_forest}).

\vspace{0.1cm}
\noindent \textbf{Correlation between sources} -  
Given between-feature correlation, the final correlation between any two sources 
$\xi_i$ and $\xi_j$ can then be estimated through
\begin{equation}
\psi(\xi_i, \xi_j) = 
\frac{1}{|\xi_i| |\xi_j|} 
\sum_{\nu \in \xi_i, \tau \in \xi_j} \lambda(\nu, \tau).
\label{eqn:src_assoc}
\end{equation}



\ifCLASSOPTIONcaptionsoff
  \newpage
\fi



\bibliographystyle{IEEEtran}
\bibliography{LearningFromMultipleSourcesForVideoSummarisation}
\end{document}